\documentclass[pdflatex,sn-mathphys-num]{sn-jnl}

\usepackage{graphicx}%
\usepackage{multirow}%
\usepackage{amsmath,amssymb,amsfonts}%
\usepackage{amsthm}%
\usepackage{mathrsfs}%
\usepackage[title]{appendix}%
\usepackage{xcolor}%
\usepackage{textcomp}%
\usepackage{manyfoot}%
\usepackage{booktabs}%
\usepackage{algorithm}%
\usepackage{algorithmicx}%
\usepackage{algpseudocode}%
\usepackage{listings}%

\algrenewcommand\algorithmicrequire{\textbf{Input:}}
\algrenewcommand\algorithmicensure{\textbf{Output:}}

\theoremstyle{thmstyleone}%
\theoremstyle{thmstyletwo}%

\theoremstyle{thmstylethree}%

\begin{document}
\title{Microstructure-based Variational Neural Networks for Robust Uncertainty Quantification in Materials Digital Twins}

\author[1]{\fnm{Andreas} \spfx{E.} \sur{Robertson}}\email{aerober@sandia.gov}

\author[1]{\fnm{Samuel} \spfx{B.} \sur{Inman}}\email{sbinma@sandia.gov}

\author[1]{\fnm{Ashley} \spfx{T.} \sur{Lenau}}\email{atlenau@sandia.gov}

\author[2]{\fnm{Ricardo} \spfx{A.} \sur{Lebensohn}}\email{lebenso@lanl.gov}

\author[1,3]{\fnm{Dongil} \sur{Shin}}\email{dongilshin@postech.ac.kr}

\author[1]{\fnm{Brad} \spfx{L.} \sur{Boyce}}\email{blboyce@sandia.gov}

\author[1]{\fnm{R\'emi} \spfx{M.} \sur{Dingreville}}\email{rdinge@sandia.gov}

\affil[1]{\orgdiv{Center for Integrated Nanotechnologies}, \orgname{Sandia National Laboratories}, \orgaddress{\state{New Mexico}, \country{USA}}}

\affil[2]{\orgdiv{Theoretical Division}, \orgname{Los Alamos National Laboratory}, \orgaddress{\state{New Mexico}, \country{USA}}}

\affil[3]{\orgdiv{Department of Mechanical Engineering}, \orgname{Pohang University of Science and Technology}, \orgaddress{\city{Pohang}, \country{South Korea}}}


\abstract{
Aleatoric uncertainties -- irremovable variability in microstructure morphology, constituent behavior, and processing conditions -- pose a major challenge to developing uncertainty-robust digital twins.
We introduce the Variational Deep Material Network (VDMN), a physics-informed surrogate model that enables efficient and probabilistic forward and inverse predictions of material behavior.
The VDMN captures microstructure‑induced variability by embedding variational distributions within its hierarchical, mechanistic architecture.
Using an analytic propagation scheme based on Taylor‑series expansion and automatic differentiation, the VDMN efficiently propagates uncertainty through the network during training and prediction.
We demonstrate its capabilities in two digital-twin-driven applications:
(1) as an uncertainty-aware materials digital twin, it predicts and experimentally validates the nonlinear mechanical variability in additively manufactured polymer composites; and
(2) as an inverse calibration engine, it disentangles and quantitatively identifies overlapping sources of uncertainty in constituent properties.
Together, these results establish the VDMN as a foundation for uncertainty‑robust materials digital twins.
}



\maketitle

\section{Introduction}\label{sec1}

Next-generation advanced manufacturing methods, such as additive manufacturing, robotic forging, and directed energy deposition, are reshaping design-for-manufacturing in low-volume, high-performance sectors (e.g., aerospace, energy).
The flexibility of these technologies surpasses that of conventional manufacturing methods, enabling the fabrication of complex geometries, localized property tailoring, and rapid design iteration.
The ability to produce bespoke components on demand significantly reduces lead times and material waste while maintaining or even enhancing performance and reliability.
However, this flexibility introduces greater variability in the material properties of the end product \cite{hu2017uncertainty}.
Variability emerges most prominently in the material microstructure, from mesoscale variations in phase or grain morphology to microscale defect distributions~\cite{dressler2019heterogeneities, roach2020size, karthik2021heterogeneous, godfrey2022heterogeneity, torquato}.
Fluctuations in processing parameters and limited controllability during manufacturing can induce significant stochastic fluctuations in mesoscale features, which in turn propagate to affect macroscopic performance characteristics~\cite{heckman2020automated, salzbrenner2017high, fernandez2022creep, karthik2021heterogeneous}.
This variability is attributable not only to the relative youth of these techniques, but also to the inherent complexity of advanced manufacturing.
Consequently, it represents \textit{aleatoric} uncertainty that cannot be fully eliminated.
To manage and mitigate such uncertainty,
developing uncertainty-robust materials digital twins -- integrating uncertainty-informed analysis \cite{hu2017uncertainty, chan2024hyperdiffusion}, 
design framework~\cite{zang2025psp, generale2024inverse}, and microstructure-aware multiscale simulation techniques such as $\mathrm{FE}^2$~\cite{tikarrouchine2018three} -- is essential for advancing and systematically deploying these advanced manufacturing technologies.


At the material length scale, developing uncertainty-robust structure–property mappings is essential for enabling these twins (e.g., in
microstructure-sensitive property estimation and design~\cite{zang2025psp, generale2024inverse},
multi-scale modeling \cite{tikarrouchine2018three},
model calibration \cite{venkatraman2022bayesian}, and
in-situ monitoring \cite{muir2021damage}).
To support these tasks, such mappings must meet three key criteria:
(1) high numerical efficiency,
(2) explicit treatment of uncertainty, and
(3) robustness across diverse use cases and input conditions.
Surrogate models are widely used due to their exceptional computational efficiency \cite{liu2019deep, shin2023deep, khatamsaz2023gpr, pasparakis2025bayesian}.
Most current approaches primarily address \textit{epistemic} uncertainty
-- uncertainty from incomplete knowledge or model inadequacy, which can, in principle, be reduced with sufficient data --
using methods such as Gaussian Process Regression \cite{khatamsaz2023gpr} and Bayesian Neural Networks \cite{pasparakis2025bayesian}.
In contrast, \textit{aleatoric} uncertainties, stemming from the inherent stochasticity of microstructural features, has only recently gained attention.
Emerging techniques, including probabilistic neural networks \cite{pourkamali2024probabilistic} and hyper-network architectures \cite{chan2024hyperdiffusion}, show promises but still lack intrinsic robustness because they omit physics-based constraints and depend heavily on training data and diversity.

The Deep Material Network (DMN) is a physics-informed machine-learning architecture designed to construct numerically efficient, robust, and deterministic structure-property relationships~\cite{liu2019deep, gajek2020micromechanics, noels2022micromechanics, shin2023deep}.
By design, DMNs are surrogate models that perform microstructural homogenization.
In a mechanical setting, they predict the homogenized stress response given
(1) a far-field strain or strain rate boundary condition and
(2) the constitutive laws of the salient microstructure features (e.g., composite phases \cite{noels2022micromechanics, shin2023deep, shin2024thermal}, or polycrystalline orientations \cite{wei2025orientation}).
Unlike black-box machine-learning models, DMNs explicitly embed microstructural physics into their architecture (Figure \ref{fig:vdmn_framework}d).
The network follows a binary-tree structure, where each parent–child building block represents a rank-1 laminate~\cite{liu2019deep, gajek2020micromechanics, noels2022micromechanics}.
Inference proceeds by recursively applying analytical homogenization through the network~\cite{gajek2020micromechanics, noels2022micromechanics, shin2023deep}.
This physics-based construction provides strong robustness and extrapolation capabilities.
DMNs generalize effectively beyond their training domain in both boundary conditions and constitutive behavior~\cite{liu2019deep, gajek2020micromechanics, noels2022micromechanics, wei2025orientation, dey2022training}.
This extrapolation capability stems directly from the underlying rank-N laminate homogenization theory~\cite{clark1994modelling, milton2022theory} (where N stands for the number of layers in the network), allowing models trained only on elastic data to extrapolate to nonlinear responses given appropriate material inputs.
Despite these advantages, current DMN implementations remain entirely deterministic and lack mechanisms for uncertainty quantification.
By construction, a DMN defines a fixed mapping from its inputs (e.g., boundary conditions, constitutive laws) to homogenized outputs.
Additionally, with few exceptions \cite{huang2022microstructurefibervariation, wei2024foundation}, microstructural information is encoded only implicitly within learned network parameters rather than explicitly represented.
As a result, existing DMNs do not explicitly capture microstructure‑induced aleatoric uncertainty, either through propagation or representation.

In this work, we introduce the Variational Deep Material Network (VDMN) as a foundation for constructing uncertainty-robust materials digital twins.
The VDMN predicts aleatoric uncertainty arising from microstructural morphology variations and supports both forward and inverse analyses of material behavior.
Unlike standard DMNs, the VDMN learns a distribution of hierarchical laminate architectures whose corresponding distribution of homogenized responses reproduces the observed microstructure-driven uncertainty.
This is accomplished by extending selected network parameters within the DMN tree architecture into hyper-variational distributions, inducing physically consistent uncertainty in the predicted outputs.
Conceptually related to the stochastic DMN of Wu and Noels \cite{wu2025stochastic}, the VDMN advances beyond prior work by introducing a general variational formulation with an autonomous training algorithm for optimizing hyper-variational parameters.
To enable such training, we developed a probabilistic homogenization scheme that analytically propagates uncertainty through the DMN architecture to predict an output distribution.
The VDMN makes the following key contributions: 

\begin{enumerate}
    \item A probabilistic propagation strategy using Taylor series expansion and automatic differentiation for analytic estimation of output uncertainty. 
    \item A log-likelihood-based training procedure that calibrates the VDMN from linear elastic data via gradient descent. 
    \item Automatic uncertainty propagation into the nonlinear regime without requiring nonlinear training data.
\end{enumerate}

We first validate VDMN training and assess its offline and online performance using a simulated dataset of phase-separated microstructures.
Our results demonstrate that the VDMN can stably extrapolate uncertainty quantification into nonlinear response regimes not represented in the training data.
We then showcase the VDMN's utility as a core component of materials digital twins through two application-driven case studies.
First, using experimental tensile samples exhibiting microstructural variability inherent to additive manufacturing, we employ the VDMN to predict (forward problem) the distribution of nonlinear mechanical responses in printed two-phase polymer composite tensile bars.
This predictive capability -- forecasting material performance prior to testing -- has the potential to reduce the cost and scope of experimental design campaigns.
Second, we apply the VDMN for inverse estimation of unknown constitutive parameters from noisy homogenized measurements, enabling quantitative analysis of multiple uncertainty sources.
We show that this uncertainty-aware calibration method can robustly disentangle uncertainty contributions from distinct aleatoric sources.
Together, these findings establish the VDMN as a robust and generalizable platform for advancing experimental analysis, design optimization, and digital twin deployment under uncertainty.

\section{Results}\label{sec2}

\begin{figure}[!ht]
    \centering
    \includegraphics[width=1.0\linewidth]{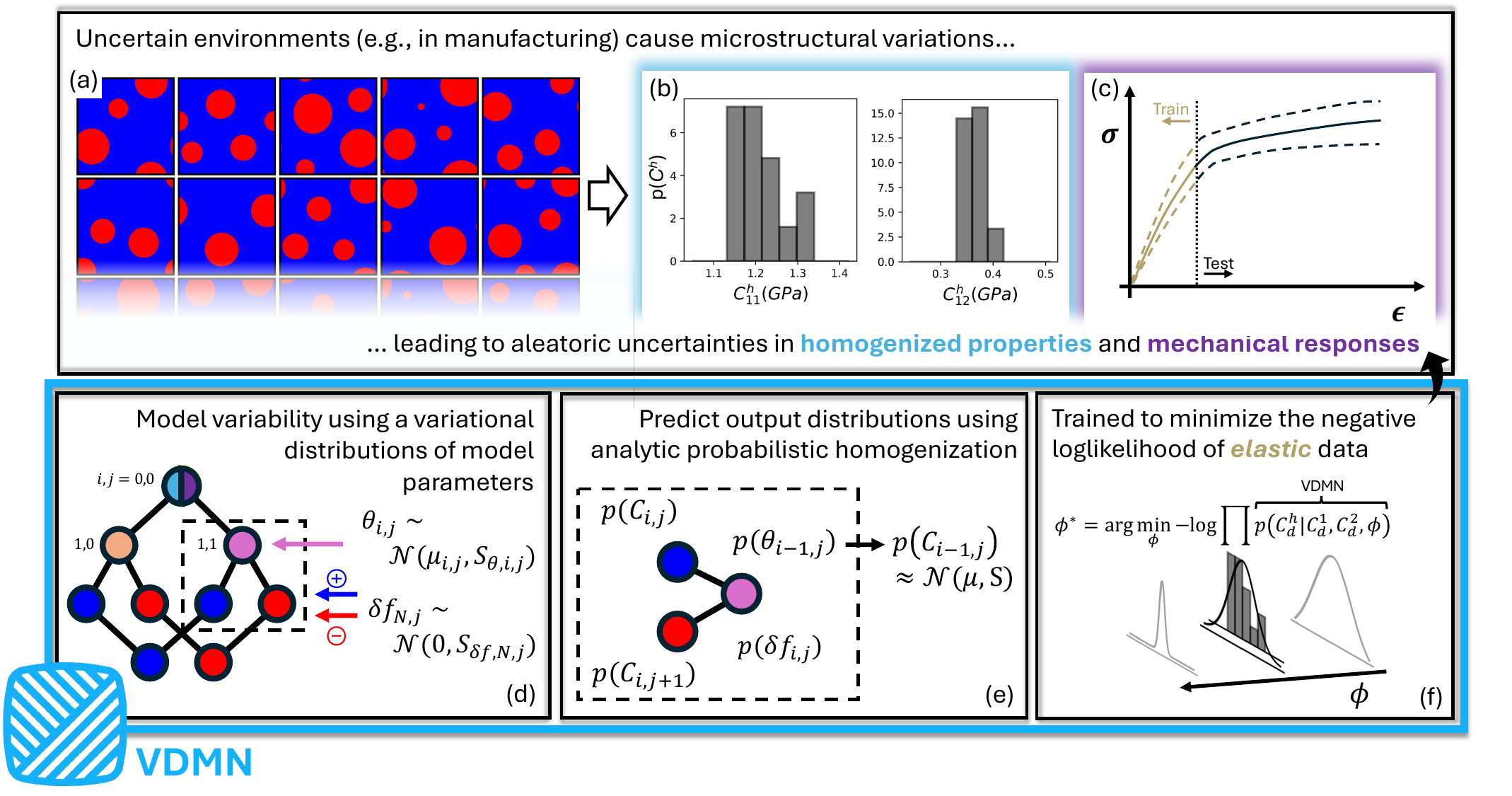}
    \caption{
    \textbf{Modeling Aleatoric Uncertainty in Materials Behavior with Variational Deep Material Networks (VDMN).}
    \textbf{Top Row: Motivation.} 
    (a) Microstructural variability in two-phase (red/blue) composites arises naturally in advanced manufacturing and represents a key source of aleatoric uncertainty. 
    (b) This variability induces distributions in homogenized elastic stiffness coefficients ($C_{11}^h, C_{12}^h$ in Voigt notation). 
    (c) Downstream, such variability propagates to variability in macroscopic linear and nonlinear responses under prescribed loading. 
    \textbf{Bottom Row: VDMN framework.} 
    (d) The VDMN augments the Deep Material Network by replacing deterministic network parameters (interface normals $\theta_{i,j}$ and volume fraction fluctuations $\delta f_{N,j}$) with variational Gaussian distributions parameterized by means $\mu_{i,j}$ and variances $S_{i,j}$. 
    Here $i$ denotes the layer index and $j$ the node index in an $N$-depth VDMN (a depth-2 example is shown). 
    (e) A probabilistic homogenization algorithm recursively propagates these uncertainties through the laminate tree to yield distributions of homogenized responses, supporting both analytic (closed-form propagation of distributions) and sampling (deterministic DMNs are drawn from the hyper-variational distributions) modes.
    (f) Network parameters $\phi$ are trained by minimizing the negative log-likelihood of elastic homogenization data. 
    Despite training on elastic responses only, the resulting model generalizes to quantify uncertainty in nonlinear regimes..
    }    
    \label{fig:vdmn_framework}
\end{figure}

VDMN extends the standard DMN by introducing hyper-variational probability distributions over the previously deterministic internal network parameters -- namely the phase volume fraction, $f$, and interface orientation, $\theta$.
In doing so, this modification transforms the DMN's hierarchical homogenization process from a fixed mapping into a probabilistic operator that outputs distributions of effective material responses (Figure \ref{fig:vdmn_framework}) rather than single deterministic values.
We formulate a stochastic homogenization algorithm that analytically propagates parameter uncertainty through the physics-constrained DMN architecture.
This homogenization algorithm combines moment estimation, Taylor expansions, and automatic differentiation, as outlined in Algorithm \ref{alg:stochastic_homogenization}.
The complete VDMN is then assembled by recursively applying this uncertainty-aware homogenization procedure through the DMN's binary-laminate tree structure, ensuring that physical constraints are maintained at every scale.
Additional theoretical background on the DMN formalism and the construction details of the VDMN are provided in the Methods section (Sec. \ref{sec:methods_mainbody}).

Conceptually, the VDMN preserves the interpretability and physics grounding of the original DMN (i.e. rank-N laminate theory) while endowing it with a generative, uncertainty-resolving capability.
The VDMN supports two complementary operational modes:
an analytic mode, in which predictive distributions are obtained from the closed-form propagation equation, and
a sampling mode, in which ensembles of deterministic DMNs are drawn from the learned hyper-variational priors.
These two modes remain consistent in their predictions; however, the sampling mode further inherits compatibility with existing nonlinear DMN implementations, \textit{now augmented with intrinsic uncertainty quantification that reflects microstructural variability}.
\begin{algorithm}[htbp] 
    \caption{\textbf{Analytic Probabilistic Homogenization within the VDMN Building Block.} This core function computes the mean and covariance of the output distribution of a single VDMN building block. In an entire VDMN, these operations are recursively combined in a binary tree structure to perform probabilistic homogenization. Einstein summation is used throughout.}
    \label{alg:stochastic_homogenization}
    \begin{algorithmic}[1] 
        \Require $\{\mu^{C^\alpha}, S^{C^\alpha}, \mu^{C^\beta}, S^{C^\beta} \}$, $\{\mu^\theta, S^\theta \}$, $\{w^\alpha, w^\beta, S^{\delta f} \}$ -- mean and covariances of input constitutive parameters ($\in \mathbb{R}^{3 \times 3}$), surface orientation ($\in \mathbb{R}^{1 \times 1}$), and child node weights and volume fraction fluctuations ($\in \mathbb{R}^{1 \times 1}$), respectively. $H(C^\alpha, C^\beta, f^\alpha, f^\beta, \theta) = C^h$ -- standard DMN homogenization function, see Sec. \ref{sec:dmn_extended_background}.
        \Ensure $\mu^{C^h}, S^{C^h}$ -- the mean and covariance of the output homogenized coefficients. 

        \State \parbox[t]{0.75\linewidth}{$H^\delta(C^\alpha, C^\beta, w^\alpha, w^\beta, \theta, \delta f) := H(C^\alpha, C^\beta, w^\alpha / (w^\alpha + w^\beta) + \delta f, w^\beta / (w^\alpha + w^\beta) - \delta f, \theta)$}
        
        \State \parbox[t]{0.75\linewidth}{$\mu^{C^h}_{ij} = H^\delta_{ij} (\mu^{C^\alpha}, \mu^{C^\beta}, w^\alpha, w^\beta, \mu^{\theta}, 0)$}
        
        \Statex \hfill \parbox[t]{0.75\linewidth}{\Comment{First-order mean: Sec. \ref{app:derivations}. Here, $i,j$ denote the indices of the homogenized stiffness tensor $\boldsymbol{C}^h$ in Voigt notation.}}
        
        \State \parbox[t]{0.75\linewidth}{$\mu_{ij}^{C^h} = \mu_{ij}^{C^h} + 0.5  \sum_{x \in \{ C^\alpha, C^\beta, \delta f \}} \\ \frac{\partial^2 H^\delta_{ij}}{\partial x_{ab} \partial x_{cd}} (\mu^{C^\alpha}, \mu^{\boldsymbol{C}^\beta}, w^\alpha, w^\beta, \mu^{\theta}, 0) S^x_{abcd}$}
        
        \Statex \hfill \parbox[t]{0.75\linewidth}{\Comment{Second-order correction. Hessian is estimated using automatic differentiation, e.g. using functorch \cite{functorch2021}.} The second order correction is often empirically unnecessary, see Supplementary Note 4 in Supplementary Information. We observe that second order $\theta$ corrections should always be avoided for stability.}

        \State \parbox[t]{0.75\linewidth}{$S^{C^h}_{ijkl}=\sum_{x \in \{ \boldsymbol{C}^\alpha, C^\beta, \theta, \delta f\}} \frac{\partial H^\delta_{ij}}{\partial x_{ab}} (\mu^{C^\alpha}, \mu^{C^\beta}, w^\alpha, w^\beta, \mu^{\theta}, 0) \\ \frac{\partial H^\delta_{kl}}{\partial x_{cd}} (\mu^{C^\alpha}, \mu^{C^\beta}, w^\alpha, w^\beta, \mu^{\theta}, 0) S^x_{abcd}$}
        
        \Statex \hfill \parbox[t]{0.75\linewidth}{\Comment{Covariance approximation under the assumption of independence among inputs, Sec. \ref{app:derivations}. Jacobians estimated with automatic differentiation.}}
        
        \State \Return $\mu^{C^h}, S^{C^h}$
    \end{algorithmic}
\end{algorithm}


The VDMN is trained using a negative log-likelihood loss (Eq. \eqref{eq:full_loss_function}), which calibrates the model to heteroskedastic uncertainty in the training data. The second term in that loss function enforces conservation of the overall phase fraction. 

\begin{equation}
    \phi^* = \arg\min_\phi \Bigg[ - \log \prod_{d=1}^D \mathcal{N} \left( \log_{\tilde{\mu}_\phi(C^{1,d}, C^{2,d})}(C^{h,d}); 0, \tilde{S}_\phi(C^{1,d},C^{2,d}) \right) + \gamma \left\|1-\sum_{i=1}^{2^l}w^i \right\|^2_2\Bigg].
    \label{eq:full_loss_function}
\end{equation}

\noindent 
Here, each data point $d$ consists of two constituent stiffness tensors, $C^{1,d}$ and $C^{2,d}$, along with the corresponding homogenized tensor $C^{h,d}$.
The VDMN predicts a mean tensor $\tilde{\mu}_\phi(C^{1,d},C^{2,d})$ and covariance tensor $\tilde{S}_\phi(C^{1,d},C^{2,d})$, which jointly define a Gaussian distribution on the tangent space of the Riemannian manifold of stiffnesses (positive semi-definite matrices).
The ground-truth homogenized tensor is projected into this tangent space via the log-map, $\log_{\tilde{\mu}_\phi}(C^{h,d})$, to evaluate the log-likelihood.
Thus, the likelihood term maximizes consistency between the predicted distribution and the observed data, while allowing for data-dependent (heteroskedastic) uncertainty.
The second term in the loss function applies a penalty to the DMN weights $w^i$, constraining their sum to unity and thereby enforcing conservation of the overall phase fraction.
This regularization preserves physical interpretability of the laminate building blocks and stabilizes the training process.
Further mathematical details are provided in Sec. \ref{sec:methods_main_vdmn}.
Here, we emphasize that this formulation enables the VDMN to reproduce observed homogenized material responses while providing calibrated, physically consistent estimates of uncertainty arising from microstructural variability.
In the following subsections, we validate these capabilities step by step:
first by demonstrating agreement between analytic and sampling modes,
then by assessing predictive performance on experimental data for two-phase composites,
and finally by evaluating inverse calibration under noisy measurement conditions.


\subsection{Validation of VDMN Predictive Performance on Synthetic Microstructure Ensembles} \label{sec:casestudy1_validatingvdmn}

Before applying the VDMN to complex experimental and inverse analyses, we first validate its predictive performance on a controlled synthetic dataset. 
Specifically, we examine whether
(i) the VDMN can be trained stably to learn accurate uncertainty representations in the offline (elastic) regime, and
(ii) this capability generalizes to online evaluations involving nonlinear material behavior. 
To this end, we constructed a benchmark dataset of $30$ stochastically varying two-phase microstructures generated via ensembles of spinodal decomposition simulations (see Supplementary Note 1 in Supplementary Information).
These simulated microstructures have comparable statistical descriptors~\cite{robertson2022efficient} but distinct morphologies, mimicking sample-to-sample variability in manufacturing processes. 
For each microstructure, we curated supervised data pairs of input constitutive parameters for the individual phases and corresponding homogenized constitutive parameters (see Supplementary Note 3 in Supplementary Information).

\begin{figure}[!ht]
    \centering
    \includegraphics[width=1.0\linewidth]{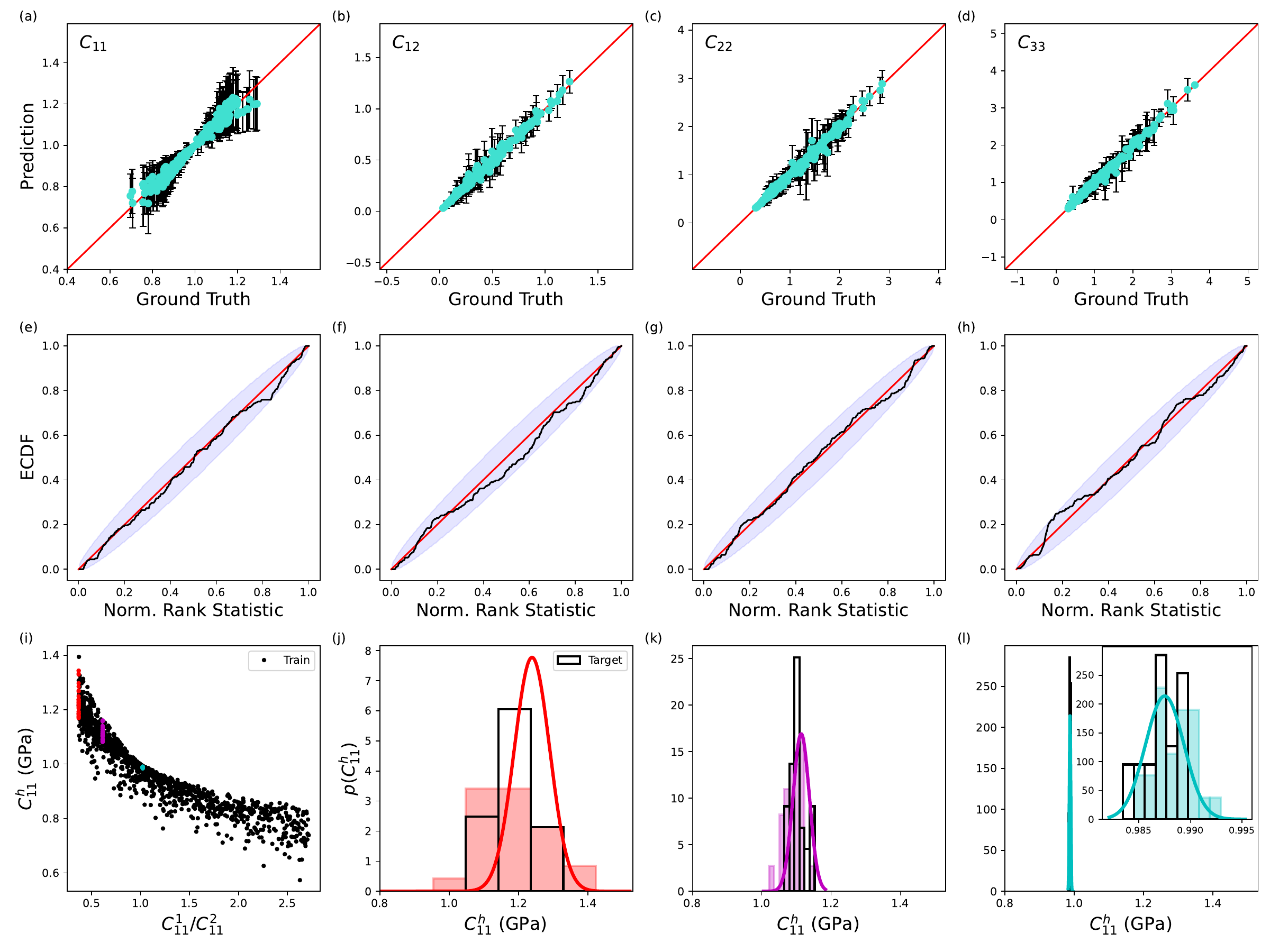}
    \caption{\textbf{Overview of Validation of Offline (Linear) Performance.}
    (a)–(d) Parity plots comparing VDMN mean and covariance predictions for four homogenized stiffness components against FFT-based ground truth; parity lines lie within three standard deviations, with marginal distributions shown for clarity. 
    (e)–(h) Simulation-based calibration tests (see Supplementary Note 2 in Supplementary Information) evaluating uncertainty prediction quality;
    black lines show normalized rank statistics relative to the VDMN Empirical Cumulative Distribution Function (ECDF),
    red lines indicate perfect calibration, and
    purple envelopes denote the 95\% confidence interval. 
    (i) Probabilistic relationship between input stiffness ratio $C^{1}_{11}/C^{2}_{11}$ and homogenized output $C^h_{11}$ for the training set (black) and three unseen test inputs, illustrating microstructure-induced variability. The three selected values were not contained in the training dataset.
    The full set of $30$ homogenized outputs -- computed on each microstructure in the ensemble -- is included for each of the three selected test inputs.
    (j)–(l) Comparison of the ground-truth output distributions (black histograms) from FFT-based homogenization over the microstructure ensemble with analytic VDMN predictions (solid lines) computed using Algorithm \ref{alg:stochastic_homogenization} and sampling-based VDMN results (colored histograms) obtained from 25 hyper-variational samples.}
    \label{fig:cs1_linearperformance}
\end{figure}

We first validate the VDMN on the probabilistic linear elastic homogenization task for which it was trained (see Supplementary Note 1 in Supplementary Information). 
Figure~\ref{fig:cs1_linearperformance} summarizes the offline performance of a 7-layer-deep VDMN evaluated on a held-out test set.
This choice of architecture (i.e number of layer) follows the configuration demonstrated to provide stable convergence and sufficient representational capacity in previous DMN studies~\cite{shin2023deep, shin2024thermomechanical}.
This depth achieves a balance between accuracy and computational efficiency, and further increases in network depth were found to yield negligible improvements in performance.
After training, the model reaches a test loss of $-50.2$.
Ablation studies examining the influence of key hyperparameters (e.g., network depth, Taylor series truncation level) are provided in Supplementary Note 4 in Supplementary Information. 
Figure~\ref{fig:cs1_linearperformance}(a)–(d) (top row) shows parity plots comparing predicted and ground-truth homogenized stiffness components across a broad range of constituent material properties.
The VDMN achieves strong performance for all components, accurately capturing both mean and variance of the data. 
A slight butterfly pattern visible in $C_{11}^h$ originates from the normalization strategy applied to the input tensors (Supplementary Note 3 in Supplementary Information) where the input $C_{11}^\alpha$ is normalized to unity as a reference.
The VDMN correctly learns that uncertainty vanishes when $C_{11}^1 = C_{11}^2$. 

The model's uncertainty quantification is statistically well calibrated across the entire dataset. 
Figure~\ref{fig:cs1_linearperformance}(e)–(h) (middle row) presents results from the simulation-based calibration test (see Supplementary Note 2 in Supplementary Information). 
Here, the black curve represents the empirical calibration curve from the test data,
while the purple envelope denotes the 95\% confidence interval. 
For all four components, the calibration curves lie well within these bounds, confirming that the predicted probability distributions are statistically consistent with the data. 

To further assess predictive distributions, we selected three representative test inputs located within the convex hull of the training set.
Ground-truth distributions were computed via numerical homogenization across the 30-microstructure ensemble for each test case.
Figure~\ref{fig:cs1_linearperformance}(i)–(l) (bottom row) demonstrates that the VDMN accurately captures the shape, mean, and variance of these distributions. 
Notably, the analytic and sampling modes of the VDMN yield nearly identical results, indicating that either mode can be used interchangeably for reliable uncertainty-aware predictions.

\begin{figure}[!ht]
    \centering
    \includegraphics[width=1.0\linewidth]{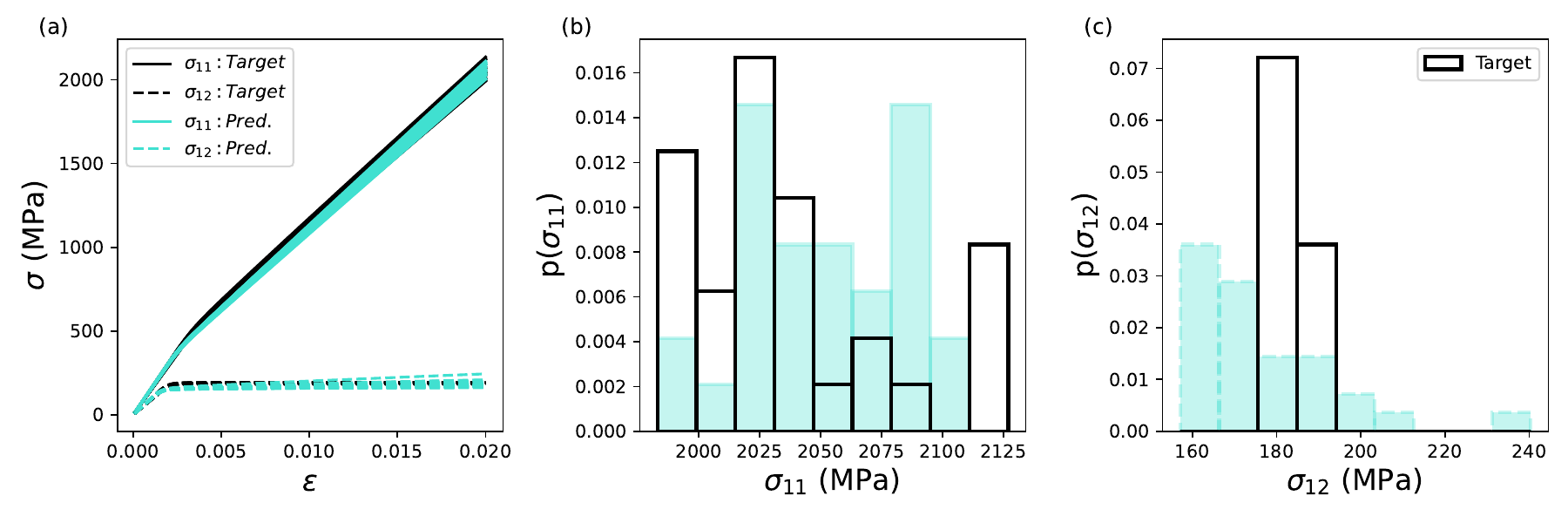}
    \caption{\textbf{Overview of Validation of Online (i.e., Non-Linear) Performance.} (a) Comparison of sample-based nonlinear VDMN predictions (i.e., predictions made by repeatedly sampling DMNs from the VDMN and running standard nonlinear predictions) against direct numerical solutions for the mechanical response of the microstructure ensemble in two different strain rate controlled boundary conditions (pure $\dot{\varepsilon}_{xx}$ and $\dot{\varepsilon}_{xy}$). (b), (c) marginal distributions comparing the sample-based predicted stress against the direct numerical solver at an applied strain of $\varepsilon_{ij}=0.02$.}
    \label{fig:cs1_nonlinearperformance}
\end{figure}

We next evaluate the VDMN's ability to extrapolate to nonlinear deformation regimes. 
Owing to attributes rooted in rank-N laminate theory, a key advantage of the DMN formalism is to make accurate nonlinear predictions without retraining, simply by substituting the linear elastic constitutive laws used during training with nonlinear ones.
The VDMN inherits this property and further extends it by enabling \textit{uncertainty quantification} in nonlinear loading scenarios -- an ability beyond the reach of conventional DMNs.
Figure~\ref{fig:cs1_nonlinearperformance} shows nonlinear predictions for a two-phase composite in which one phase is linear elastic ($E=100{,}000$~MPa, $\nu=0.3$) and the other follows a Norton elastoviscoplastic law ($E=200{,}000$~MPa, $\nu=0.19$, $\sigma_y=\sigma_y^\mathrm{max}=300$~MPa, $\delta=0$, $K_p=0$, $N=10$; see Supplementary Note 1 in Supplementary Information). 
The composite microstructure is subjected to pure strain-rate boundary conditions up to $\varepsilon=0.02$, and the VDMN is used in online mode (Sec.~\ref{sec:vdmn_online_mode}) to predict the corresponding (nonlinear) stress–strain responses.

A major benefit of the online mode is its computational efficiency:
each nonlinear VDMN inference requires approximately $2.7$ CPU-seconds in serial execution,
compared to $\sim 86{,}400$ CPU-seconds for the equivalent full-field numerical simulation.
In addition to accurately reproducing the mean response, the VDMN predicts the full distribution of mechanical responses, thereby quantifying the aleatoric uncertainty induced by microstructural variability --  a capability unavailable with standard DMNs.
Figures~\ref{fig:cs1_nonlinearperformance}(b),(c) compare predicted marginal distributions at $\varepsilon=0.02$ under tensile and shear loading, showing good agreement with the reference data, particularly in tension -- consistent with previous DMN analyses~\cite{shin2023deep}.

\subsection{Applications} \label{sec:casestudy2_applications}

We demonstrate the utility of the VDMN as a foundational uncertainty-aware materials digital twins.
Acting as a distribution of surrogate microstructures, the VDMN enables the rapid and systematic exploration of how microstructural variability influences macroscopic material behavior.
Here, we apply the VDMN to experimental datasets to analyze uncertainty arising from manufacturing-induced heterogeneities.
Specifically, we address two critical tasks:
(i) forward uncertainty quantification, where the VDMN predicts distributions of nonlinear mechanical responses of fabricated test samples, and
(ii) inverse uncertainty quantification, where it performs uncertainty-robust calibration of constitutive parameters from noisy experimental measurements. 
Additional experimental and computational details are provided in Supplementary Information.

\subsubsection{Forward Uncertainty Quantification on 3D Printed Samples}  \label{sec:casestudy2_3D_Print}

\begin{figure}[!ht]
    \centering
    \includegraphics[width=1.0\linewidth]{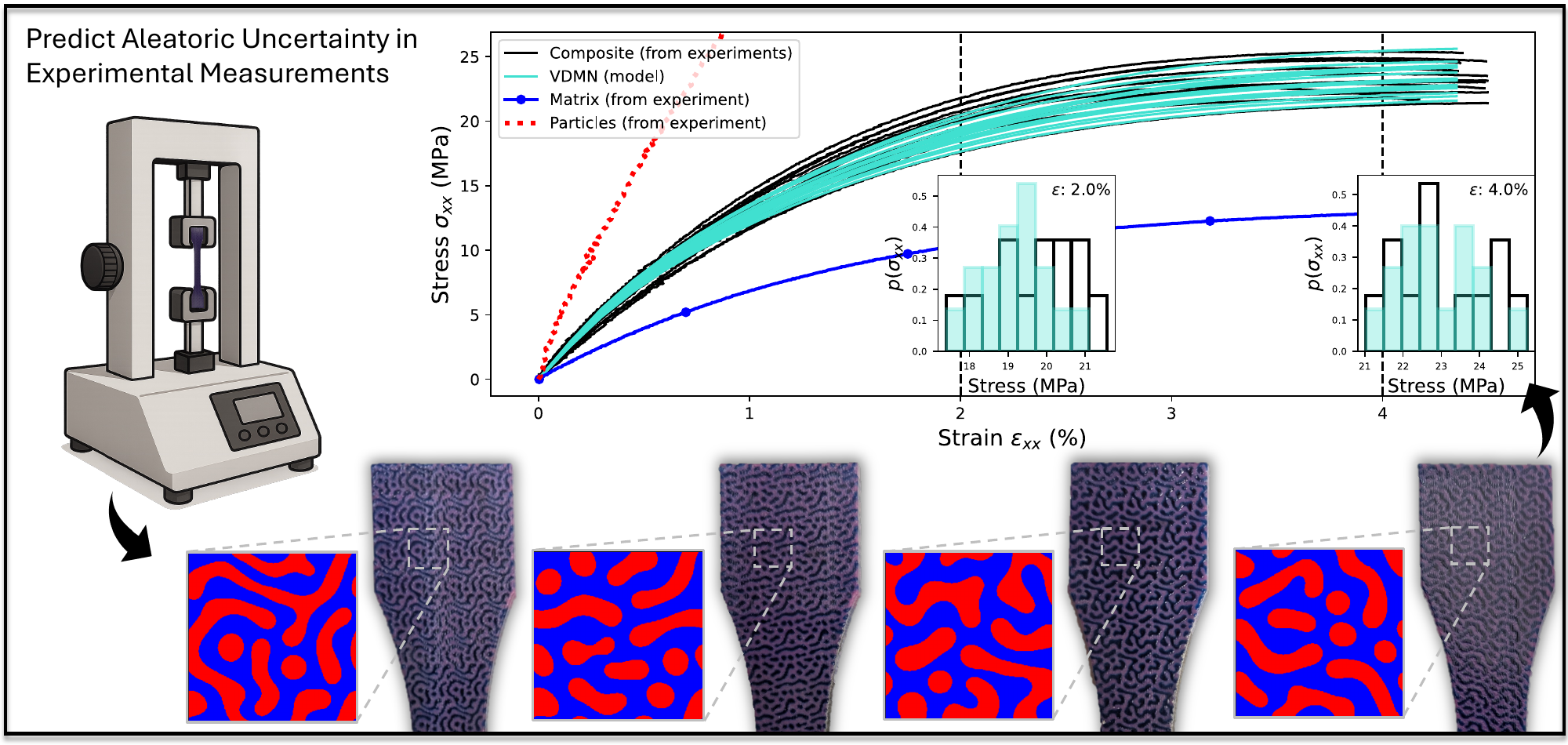}
    \caption{\textbf{Comparison of VDMN Predictions with Experimental Tensile Measurements on 3D Printed Two-Phase Composites.}
    Four representative tensile bars (from twelve) fabricated via polymer additive manufacturing are shown, each derived from a distinct spinodal unit cell to emulate microstructural variability.
    For each specimen, the base unit cell and a segment of the printed bar are displayed.
    Insets show VDMN ensemble predictions (turquoise) of tensile stress-strain responses, compared with experimental measurements (black).
    The experimentally measured stress-strain responses of the matrix phase (blue) and particle phase (red) are also depicted.
    These measurements were performed prior to the VDMN predictions and used to calibrate the Norton elastoviscoplastic constitutive models inputted to the VDMN, see Supplementary Information.
    Marginal stress distributions at two strain levels highlight agreement between predicted and measured variability.}
    \label{fig:cs2_forwardmodeling}
\end{figure}

Microstructural uncertainties naturally arise during manufacturing as constitutively simple materials develop complex microstructure morphologies.
For downstream tasks such as design~\cite{hu2017uncertainty}, quantifying uncertainty caused by these variations without relying on time-consuming experimental testing campaigns is desirable.
The VDMN directly enables this by predicting mechanical response variability induced by stochastic differences across manufactured microstructure ensembles.
We demonstrate this capability through an experimental case study using polymer additive manufacturing.
We fabricated an ensemble of tensile specimens with varied spinodal unit cells (Figure~\ref{fig:cs2_forwardmodeling} and Supplementary Note 5 in Supplementary Information).
Each printed specimen consists of two interwoven elastoviscoplastic phases, whose effective mechanical behavior reflects both intrinsic material properties of the polymer phases and microstructural morphology.
Constitutive parameters for individual phases were independently calibrated from single‑phase tests, while parameter for the composites remained unknown.
Using only the individual phase properties and the prescribed microstructural variability (e.g., from optical microscopy), the VDMN accurately predicts both the homogenized mechanical response and its associated uncertainty (Figure~\ref{fig:cs2_forwardmodeling}).
Inset panels report close agreement between predicted and experimental stress–strain distributions, including at large strains.
Remarkably, the VDMN correctly captures nonlinear response uncertainty despite training solely on inexpensive linear-elastic data, maintaining strong performance on complex spinodal microstructures typically challenging for standard DMN implementations~ \cite{shin2023deep}.
Simulations matched experimental mixed boundary conditions (see Supplementary Note 5 in Supplementary Information), preserving predictive fidelity.

This comparison underscores the VDMN's strength as an efficient and predictive materials digital twin.
Unlike conventional deterministic DMNs that reproduce only the nominal and average responses,  the VDMN inherently captures variability from microstructural randomness via its hyper‑variational formulation.
It predicts full uncertainty distributions rather than single outcomes, enabling realistic, uncertainty-aware digital material representations.
Equally important, the VDMN combines this capability with exceptional computational and data efficiency.
Trained on inexpensive linear-elastic data, it generalizes to nonlinear, complex regimes without retraining or fine‑tuning.
Thus, it achieves distributional fidelity comparable to brute-force direct numerical simulations (DNS) at a fraction of the cost, offering a practical, scalable surrogate for real-time materials analysis, design, and digital twin deployment.

\subsubsection{Inverse Uncertainty Quantification}  \label{sec:casestudy3_inverseUQ}

\begin{figure}[!ht]
    \centering
    \includegraphics[width=1.0\linewidth]{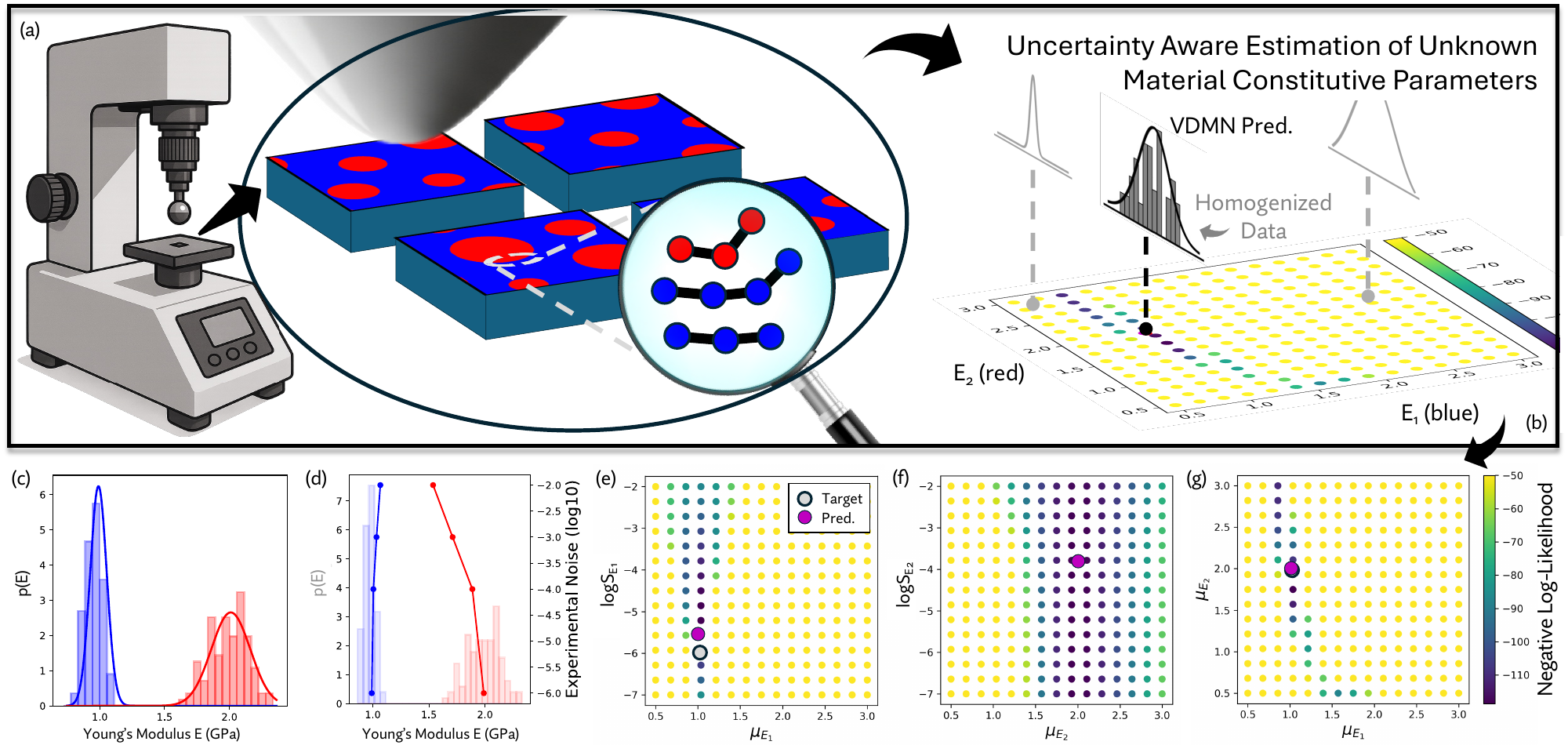}
    \caption{\textbf{Uncertainty-Aware Calibration of Constitutive Properties.} 
    (a) Three primary sources of uncertainty (experimental, mesoscale microstructural, and microscale constitutive uncertainty) arise during mechanical testing (indentation shown as an example). 
    The underlying constitutive property distributions are often difficult to measure directly. 
    (b) The VDMN disentangles these uncertainty sources and estimates unknown constitutive distributions by identifying parameters that maximize the likelihood of observed homogenized measurements under a VDMN-based total probability model (see Supplementary Note 6 in Supplementary Information). 
    (c) Comparison between predicted constitutive distributions (solid lines) and ground truth (histograms) for two isotropic phases where only the Young's moduli $E$ vary (constant Poisson's ratio, $\nu=0.3$). 
    (d) Effect of increasing experimental noise (variance) on the predicted mean Young's modulus $E$ for each phase.
    (e–g) Negative log-likelihood landscapes of homogenized measurements with respect to unknown constitutive parameters, computed using the VDMN total probability model. 
    Phase~1 corresponds to the matrix phase (blue). 
    These landscapes quantify parameter confidence and guide experimental design (see Supplementary Note 6 in Supplementary Information).}
    \label{fig:cs3_inversemodeling}
\end{figure}

In the previous experimental case, the sources of aleatoric uncertainty, i.e. the microstructure variability, were known or easily measurable.
However, in many practical settings, some input uncertainties are not directly measurable.
Instead, these latent distributions must be inferred from observed variability in aggregate experimental measurements.
Figure~\ref{fig:cs3_inversemodeling} illustrates a representative scenario in which the homogenized elastic coefficients of multiple samples, produced under identical processing conditions, exhibit statistical variation.
The measured variability reflects the combined influence of
(1) microstructural uncertainty,
(2) measurement noise, and
(3) uncertainty in the unknown constitutive properties of the material phases.
While the first two can be approximated (e.g., via imaging and repeated testing), the third (for instance due to aging of printing powder) is generally inaccessible and must be estimated.
The VDMN provides this estimation capability by serving as a probabilistic model for inverse uncertainty quantification.

We constructed a full approximation of the uncertainty in this system utilizing the stochastic propagation framework introduced described in Sec. \ref{app:derivations}.
We propagated uncertainties in the constitutive properties of each phase through the VDMN, while using an additive Gaussian noise model to capture experimental noise (see Supplementary Note 6 in Supplementary Information).
Assuming Gaussian distributions for the unknown phase constitutive parameters, we identified their means and variances by maximizing the likelihood of the measured homogenized data using the VDMN-based total uncertainty model. 

To validate this approach, we performed a controlled simulation study. 
Assuming isotropic behavior for both phases, we prescribed distributions on the Young's moduli ($E$) of each phase, while Poisson's ratios were fixed at $0.3$.
Homogenized measurements were generated using a Fast Fourier Transform (FFT)-based direct numerical solver (see Supplementary Note 6 in Supplementary Information).
As shown in Figure \ref{fig:cs3_inversemodeling}(c), the VDMN-recovered constitutive distributions (solid lines) closely match the ground-truth input distributions (histograms), achieving relative absolute errors of $1\%$, $8.1\%$, $0.5\%$, and $<0.1\%$ for the mean and log-variance of the Young's moduli of each phase ($\mu_{E_1}$, $\log S_{E_1}$, $\mu_{E_2}$, and $\log S_{E_2}$, respectively; see Supplementary Note 6 in Supplementary Information).
This excellent agreement, obtained in the absence of experimental noise, clearly demonstrates the VDMN's ability to accurately infer underlying constitutive variability.

When experimental noise is introduced (Figure \ref{fig:cs3_inversemodeling}(d)), recovering the underlying input constitutive variability becomes increasingly difficult. 
Model predictions remain robust until the magnitude of measurement noise approaches that of the constitutive uncertainty.
This breakdown can be understood by examining the likelihood landscape of the homogenized measurements as a function of the input distribution parameters (Figure~\ref{fig:cs3_inversemodeling}(e-g)).
Although the optimized solution coincides with the ground truth, multiple parameter combinations yield nearly identical likelihoods.
Such degeneracy indicates parameter sensitivity within the VDMN optimization and quantifies the epistemic uncertainty of the total uncertainty model.
Highly sensitive parameters, such as $\mu_{E_1}$, can still be inferred reliably even under significant measurement noise.
Moreover, these likelihood landscapes can guide experimental design at the homogenized scale by revealing which homogenized measurements most effectively constrain the underlying constitutive distributions.
In this case study, we found that measuring $C_{11}$ and $C_{12}$ provided the highest information value for parameter recovery, as shown in Figures~\ref{fig:cs3_inversemodeling}(e)-(g) (see Supplementary Note 6 in Supplementary Information).

Finally, the complexity of the inverse problem depends on both phase property contrast and the degree of microstructural variability.
Large contrasts or dominant uncertainty sources simplify identification, whereas balanced contributions complicate disentanglement.
Despite this challenge, the VDMN consistently resolves the true constitutive distributions, demonstrating its robustness and efficiency for inverse uncertainty quantification in materials systems with intertwined sources of variability.

\section{Discussion}
\label{sec:discussion}

Next-generation manufacturing will increasingly rely on materials digital twins for design analysis, predictive monitoring, process control, and closed‑loop decision‑making~\cite{kalidindi2022digital}.
As demonstrated above, the VDMN provides a powerful foundation for such digital twins by serving as an efficient, physically grounded, and uncertainty‑aware surrogate microstructure.
Based on rank-N laminate homogeniztion theory, the VDMN integrates physics‑based interpretability with machine‑learning flexibility, enabling stable, accurate prediction of homogenized material behavior across diverse loading conditions, processing routes, and constitutive behaviors.
Its microstructure‑informed, sparse hierarchical graph architecture makes inference speeds orders of magnitude faster than RVE-based digital twins, facilitating real‑time deployment in manufacturing digital twins.
Additionally, the Taylor‑series‑based analytic propagation of variational distributions rigorously models aleatoric uncertainty from microstructural variability.

Our results demonstrate that the VDMN fulfills core materials digital twin functions:
(1) extrapolating predictive uncertainty across different constitutive behaviors and loading conditions,
(2) performing forward uncertainty quantification for manufactured parts, and
(3) enabling inverse calibration of unknown material parameters despite noisy or incomplete data.
Importantly, these capabilities extend beyond synthetic testbeds to experimental settings, confirming robustness under realistic manufacturing conditions.
This high fidelity is achieved with minimal training cost from inexpensive linear‑elastic data without retraining for nonlinear responses.

Looking forward, the VDMN framework offers a scalable foundation for multimodal, multiphysics digital twins~\cite{boyce2023machine, walker2024unsupervised}.
Its rank-N laminate theoretical basis supports extensions to other physical phenomena~\cite{shin2024thermal} and material classes~\cite{wei2025orientation}.
Integrating Bayesian and active-learning pipelines would enable simultaneous treatment of epistemic and aleatoric uncertainties~\cite{chan2024hyperdiffusion, kennedy2001bayesian}, allowing digital twins to update and self‑calibrate as new as new data arrive.
Beyond uncertainty quantification, the VDMN could serve as a decision‑support layer within intelligent manufacturing systems, providing real-time feedback to adaptive control, guiding materials-by-design workflows, or acting as a surrogate in reinforcement-learning-based process optimization~\cite{desai2022learning}.
Coupled with in-situ sensing and edge computing, such models could close the loop between experiment, simulation, and control, advancing materials digital twins toward autonomous operation~\cite{kalidindi2022digital, szymanski2023autonomous, boyce2023machine}.

\section{Methods}\label{sec:methods_mainbody}

This section a concise overview of the core components of the VDMN.
Comprehensive methodological details, including full algorithmic descriptions, ablation studies, and discussions of existing methods, are provided in the accompanying Supplementary Information, along with expanded presentations of the two application areas: forward and inverse uncertainty quantification.

\subsection{Background}
\label{sec:methods_mainbody_background}

\subsubsection{Deep Material Network Background}
\label{sec:dmn_extended_background}

Based on rank-N laminate theory~\cite{clark1994modelling,milton2022theory}, the Deep Material Network (DMN) framework predicts the homogenized response of a material microstructure by identifying an equivalent surrogate microstructure~\cite{liu2019deep, gajek2020micromechanics, noels2022micromechanics, shin2023deep}.
This surrogate microstructure takes the form of a graphical (often, binary) tree in which the connection between child and parent nodes represents a homogenization operation grounded in the theory of composites~\cite{clark1994modelling,milton2022theory} (Figure~\ref{fig:vdmn_framework}, bottom left).
Specifically, each child-parent triplet (the black dashed box) represents a two-phase laminate microstructure.
Different DMN architectures can be constructed depending on the form of this fundamental laminate; the only constraint is that its homogenized properties must be computable analytically.
In this work, we use the interaction based DMN construction outlined by Shin \textit{et al.} \cite{shin2023deep}.
In this formulation, each laminate is characterized by:
(1) a unit vector defining the laminate orientation of the interface between the two phases represented by the child nodes, represented by an angle $\theta_{i,j}$ (in our notation, the first subscript refers to the layer and the second refers to the node within the layer, see Figure \ref{fig:vdmn_framework}) for 2D laminates, and
(2) a weight associated with each child node, $w_{i+1,2j}$ and $w_{i+1,2j+1}$.
The volume fraction of each child phase, required for homogenization, is computed through weight normalization: $f_{i+1,2j} = w_{i+1,2j}/(w_{i+1, 2j} + w_{i+1, 2j+1})$.
To reduce the number of free parameters, parent weights are inherited as the sum of their child weights, $w_{i,j} = w_{i+1, 2j} + w_{i+1, 2j+1}$.
The definition of parent and child node has been adopted to follow the original work of Liu and coworkers~\cite{liu2019deep}.
With this weight‑tying scheme, a N-layer DMN representing a 2D two-pahse microstructure contains $N_w = 2^N$ weight parameters and $N_\theta = 2^N-1$ orientation parameters (e.g., $N_w=4$ and $N_\theta=3$ in the $2$ layer DMN depicted in Figure \ref{fig:vdmn_framework}).

The DMN operates in two principal modes: offline and online.
In the offline mode, the DMN functions as an elastic homogenizer that maps the elastic stiffness tensors of its constituent phases to the effective (homogenized) stiffness tensor of the composite.
Homogenization is performed through a hierarchical sequence of elemental unit operations corresponding to the network's graphical tree structure (Figure \ref{fig:vdmn_framework}).
In a DMN for a two-phase composite, each fundamental building block comprises two child nodes, $n_{i+1,2j}$ and $n_{i+1, 2j+1}$, and one parent node, $n_{i,j}$.
The children provide their constitutive tensors $C_{i,j}$, associated weights, $w_{i,j}$, and an interface orientation $\theta_{i,j}$.
The parent node's homogenized stiffness tensor is then computed analytically from these quantities according to the unit‑homogenization relation described below.

\begin{equation}
\begin{split}
    C^h_{i,j} = &f_{i+1, 2j}C_{i+1, 2j} \left( 1 + \frac{1}{f_{i+1, 2j}} H(\theta_{i,j}) B_{i,j} \right) + \\
    &f_{i+1, 2j+1}C_{i+1, 2j+1} \left( 1 - \frac{1}{f_{i+1, 2j+1}} H(\theta_{i,j}) B_{i,j} \right).
\end{split}
\label{eq:deterministic_homogenization_process}
\end{equation}

\noindent Here, $f_{i+1, 2j}$ is the volume fraction which is computed from the child weights: $f_{i+1, a} = w_{i+1, a} / (w_{i+1, 2j} + w_{i+1, 2j+1})$ for $a=\{2j, 2j+1 \}$ and $H(\cdot)$ is the orientation matrix. 

\begin{equation}
    H(\theta) = \left[ 
    \begin{array}{ccc}
        \cos{2\pi\theta} & 0 & \sin{2\pi\theta} \\
        0 & \sin{2\pi\theta} & \cos{2\pi\theta}
    \end{array} \right]^T.
\end{equation}

\noindent $B_{i,j}$ is defined as follows. 


\begin{equation}
\begin{split}
B_{i,j} &= -f_{i+1,2j}f_{i+1, 2j+1} \\
&\quad \times \left[H(\theta_{i,j})^T(f_{i+1,2j}C_{i+1,2j+1} + f_{i+1,2j+1} C_{i+1,2j}) H(\theta_{i,j}) \right]^{-1} \\
&\quad \times H(\theta_{i,j})^T(C_{i+1,2j} - C_{i+1,2j+1}).
\end{split}
\end{equation}

In the online mode, the DMN functions analogously to a traditional finite‑element–based microstructural simulation.
An applied far‑field boundary condition (typically a strain‑rate tensor) is propagated through the network, distributing local strain‑rate fields to each node.
The corresponding local stresses are then evaluated using standard constitutive relations (see Supplementary Note 1 in Supplementary Information).
In this work, we adopted Norton elastoviscoplasticity for the phase constitutive behavior.
Once local stresses are obtained, they are recursively homogenized through the DMN's hierarchical architecture to yield the overall (homogenized) stress response.
The homogenization procedure introduces an intermediate quantity known as the jumping vector, which enforces equilibrium across laminate interfaces.
This vector is determined iteratively using a Newton-Raphson update scheme. 
Compared with direct numerical simulations (DNS) based on representative volume elements (RVEs), the DMN requires significantly fewer degrees of freedom since its surrogate microstructure captures the essential mechanics in a reduced‑order form. 
Consequently, the computational cost for online evaluations is reduced by several orders of magnitude.
In this work, we employed the implementation described \cite{shin2023deep} without modifications.
A detailed formulation of the base DMN architecture, including its 3D extensions and implementation details, can be found in Shin et al.~\cite{shin2023deep}.

\subsubsection{Riemannian Statistics}
\label{sec:reimannian_statistics}

We utilized Riemannian statistics to represent probability distributions over objects that exist on Riemannian manifolds \cite{calinon2020reimanniangaussians, pennec2006intrinsicreimannian}.
A Riemannian Gaussian is defined as a multivariate Gaussian distribution formulated in the tangent space constructed with respect to the distribution's mean.
The tangent space provides a locally Euclidean (vectorized) approximation of the manifold, enabling the use of standard Gaussian densities without explicitly accounting for the Gaussian's unbounded support.
Although, in theory, the tangent space is bounded by the reference point's cut lotus -- the set of points with a non-unique minimizing shortest line (geodesic) back to the original point -- it is common in Riemannian statistics to assume that the mean lies sufficiently far from this region such that the tangent space can be treated as effectively unconstrained.

In this work, we are specifically interested in Riemannian Gaussian distributions defined over the space of Positive Semi-Definite (PSD) matrices (Sec. \ref{sec:methods_main_vdmn}).
We constructed this distribution as a zero-mean multivariate Gaussian distribution with a prescribed covariance matrix centered at the center of the tangent space.
The probability density of an observed PSD matrix was then computed by mapping it into this tangent space using the analytic logarithmic map for PSD matrices (Eq. \eqref{eq:app_logarthmicmap_PSD}).
For a given mean PSD matrix, $A$, the logarthmic map is defined by the follow expression.

\begin{align}
    C &= \log_A (B) \\
    &= A^{1/2} \log \left( A^{-1/2} B A^{-1/2} \right) A^{1/2}.
    \label{eq:app_logarthmicmap_PSD}
\end{align}

\noindent This mapping transforms an arbitrary matrix, $B$, into an element of the tangent space of PSD matrices around $A$, $C$.
In our formulation,
$A$ corresponds to the mean homogenized stiffness matrix (in Voigt notation) predicted by the VDMN,
$B$ is the target stiffness matrix, and
$C$ is its tangent‑space projection.
The resulting predictive density is given by Eq.~\eqref{eq:variationalapproximation}.

\subsection{Variational Deep Material Network}
\label{sec:methods_main_vdmn}

The VDMN  extends the standard DMN framework to model the homogenization uncertainty arising from irreducible microstructural variability.
In the deterministic DMN, the optimized network topology is often interpreted as a surrogate representation of the underlying microstructure morphology~\cite{liu2019deep, gajek2020micromechanics, shin2023deep, noels2022micromechanics}.
The VDMN generalizes this concept by treating an ensemble of network topologies as a probabilistic representation of a microstructure ensemble, thereby enabling direct modeling of aleatoric uncertainty.
The proposed VDMN introduces a variational parameterization by replacing selected deterministic parameters in the DMN architecture with Gaussian distributions.
These variational parameters collectively define the distribution that governs the surrogate microstructure ensemble (Figure \ref{fig:vdmn_framework}).
Specifically, the laminate orientation angles, $\theta$, are replaced with learned hyperparameters, $\mu_\theta$ and $S_\theta$, representing the mean and variance of their Gaussian distributions.
These orientation distributions are defined at every layer of the network.
To further capture stochastic variability in phase volume fractions at the finest scale, the parent nodes in the input layer of the VDMN are assigned a volume fraction perturbation, $\delta f$, modeled as a zero-mean Gaussian with trainable variance, $S_{\delta f}$. 
This perturbation is applied by adding $\delta f$ to one child phase and subtracting it from the other, preserving total volume fraction while introducing controlled variability.
The perturbation is restricted to the input layer to prevent non‑linear coupling effects in the stress homogenization term (see Eq\@.~(16) in~\cite{shin2023deep}).
From this formulation, the VDMN now outputs a probability distribution rather than a single deterministic prediction.
In the offline setting, the output is the following conditional distribution.

\begin{equation}
    v_\phi : \boldsymbol{C}^\alpha, \boldsymbol{C}^\beta \mapsto p(\boldsymbol{C}^h | \boldsymbol{C}^\alpha, \boldsymbol{C}^\beta;\phi),
    \label{eq:vdmn_abstract_function}
\end{equation}
\noindent Here, $v_\phi$ is the VDMN with trainable parameters $\phi$, $\boldsymbol{C}^\alpha$ and $\boldsymbol{C}^\beta$ denote the stiffness tensors of the constituent phases, and $p(\boldsymbol{C}^h)$ represents the predicted homogenized stiffness distribution.

\subsubsection{Loss Functions for the VDMN}

Transitioning to a probabilistic formulation necessitates defining a new loss function.
Several loss functions are commonly employed in variational settings, including
divergence based methods \cite{neal1992bayesian, daubener2025elbo},
maximum mean discrepancy distances \cite{li2015generative, borgwardt2006integrating}, and
negative log-likelihoods \cite{papamakarios2021normalizing, pourkamali2024probabilistic, sun2022alpha, daubener2025elbo}.
Amongst these, the negative log-likelihood formulation is particularly well-suited to conditional regression problems such as ours, where the goal is to learn a conditional distribution of homogenized stiffness given phase properties. 
Accordingly, the primary component of the loss is defined as

\begin{equation}
    L(\phi, \{\boldsymbol{C}^{1, b}, \boldsymbol{C}^{2,b}, \boldsymbol{C}^{*,h,b} \}_{b \in B}) = -\sum_{b \in B} \log p(\boldsymbol{C}^{*,h,b} | \boldsymbol{C}^{1,b}, \boldsymbol{C}^{2,b}, \phi),
\end{equation}

\noindent where $\phi$ is the set of trainable parameters (i.e., the parameters of the hypervariational distributions and the weights at the base nodes) and
$B$ represents the batch used for training.
We assumed that all the observed training data is independent and identically distributed.
Although compact in form, the expression above is actually deceptively computationally complex.
Specifically, to evaluating the loss, we require a closed form expression for the predicted conditional distribution in order to calculate the loss for a given mini-batch during training.
This is particularly challenging because the VDMN homogenization functions are nonlinear.
The complete loss function, provided in Eq\@.~\eqref{eq:full_loss_function} also includes the traditional penalization on the sum of the weights (e.g., Eq. (21) in \cite{shin2023deep}) to ensure numerical stability and physical consistency.

\subsubsection{Approximating the Output Density}

We approximate the output density of the VDMN as a Riemannian Gaussian distribution~\cite{calinon2020reimanniangaussians, pennec2006intrinsicreimannian}.
The theoretical basis for this choice and its implications are discussed in Sec. \ref{sec:reimannian_statistics}, with further practical evaluations presented in the ablation study in Supplementary Note 4 in Supplementary Information.
For the current context, it suffices to note that constructing this approximation requires estimates of the mean and covariance of the predicted distribution.
These statistical moments are computed using the method introduced by Venkatraman \textit{et al.} \cite{venkatraman2025matse}.

Briefly, we consider a single homogenization unit within the VDMN (e.g., the black dashed box in Figure \ref{fig:vdmn_framework}).
During offline training, the parent node receives as input the weights and constitutive properties of its two child nodes.
Within the VDMN, the input constitutive coefficients, except for the input layer, are random variables, reflecting uncertainty propagated from lower layers.
The weights, by contrast, remain deterministic due to the imposed zero‑mean condition of the volume‑fraction variations.
The parent node also incorporates its local laminate orientation (the normal vector) and, at the input layer, its own stochastic volume‑fraction perturbation.
The node outputs the homogenized constitutive coefficients, which then propagate upward through the network.
We emphasize here that only the variability in these coefficients is propagated upwards across layers, ensuring that uncertainty is consistently carried through the hierarchical structure.

Given the first‑ and second‑order moments of the random inputs and a deterministic homogenization operator (Eq\@.~\eqref{eq:deterministic_homogenization_process}), the target is to derive analytic estimates of the corresponding first and second moments of the parent's output coefficients.
We adapted  the formulations presented in Venkatraman \textit{et al.} to remove uncertainty in the function itself \cite{venkatraman2025matse} and extend to tensor-valued (e.g., matrix) functions.
The full derivation is provided in Sec. \ref{app:derivations}.

\begin{equation}
    \mathbb{E}_{p(\boldsymbol{x})}[f_{ij}(\boldsymbol{x})] \approx  f_{ij}(\bar{\boldsymbol{x}}) + 
    \frac{1}{2!} \frac{\partial^2 f_{ij}}{\partial x_a \partial x_b}\bigr|_{\boldsymbol{x} = \bar{\boldsymbol{x}}} S_{ab}. \\
    \label{eq:mean_estimate}
\end{equation}

\noindent Here, $\mathbb{E}[\cdot]$ is the expectation operator.
For notational compactness, the VDMN's inputs are assumed to be concatenated and vectorized, $\boldsymbol{x}$.
This expression is easily generalizable to tensors of arbitrary order; in practice,  we found implementation easier in an expanded form for matrix equations.
The vector $\bar{\boldsymbol{x}}$ denotes the mean of the input distribution, and $S_{ab}$ is the covariance between the $a$ and $b$ components.
Einstein summation convention is adopted throughout, such that repeated indices imply summation.
We assumed statistical independence for the covariance between any pair of the inputs (i.e., zero inter‑variable covariance), while allowing each input to possess an arbitrary internal covariance structure.
In practice, we found that retaining only the first‑order mean term—omitting the second‑order (Hessian) correction—was sufficient to ensure stable VDMN training (see Supplementary Note 4 in Supplementary Information). Under this approximation, the output covariance can be estimated as,

\begin{equation}
    \textrm{COV}[\boldsymbol{f}(\boldsymbol{x})]_{ijkl} \approx  \frac{\partial f_{ij}}{\partial x_a}\bigr|_{\boldsymbol{x} = \bar{\boldsymbol{x}}} \frac{\partial f_{kl}}{\partial x_b}\bigr|_{\boldsymbol{x} = \bar{\boldsymbol{x}}} S_{ab}.
    \label{eq:variance_estimate}
\end{equation}

\noindent Here, $\textrm{COV}[\cdot]$ is the covariance operator, and all derivatives are computed automatically via automatic differentiation~\cite{functorch2021}.
These operations define how uncertainty propagates through each individual unit of the VDMN (Algorithm \ref{alg:stochastic_homogenization}).
The full network is then assembled by connecting these building blocks in the standard binary tree architecture. 
Each unit operates independently of the others at the same layer.
Therefore, the dimensionality of the expectations and derivatives remains invariant to network depth or size, a property that highlights the computational efficiency and scalability of the VDMN's hierarchical binary structure. 

\subsubsection{Output Densities of Coefficient Matrices}

Stiffness matrices -- i.e., stiffness tensors represented in Voigt notation -- are constrained to be positive semi‑definite (PSD) in accordance with Drucker stability.
As such, these matrices reside on a Riemannian manifold, specifically the PSD cone.
To account for this geometric structure, we employ Riemannian statistics rather than standard Euclidean formulations (Section \ref{sec:reimannian_statistics}).
We define the Riemannian Gaussian in the tangent space constructed around the VDMN's mean prediction.
Using the VDMN‑predicted mean and covariance, together with the standard multivariate Gaussian distribution and the logarithmic map on the PSD manifold (Eq.~\eqref{eq:app_logarthmicmap_PSD}), the probability density of a predicted stiffness matrix is expressed as

\begin{equation}
    p(\boldsymbol{C}^h | \boldsymbol{C}^1, \boldsymbol{C}^2, \phi) \approx \mathcal{N} \left(\log_{\tilde{\mu}_\phi(\boldsymbol{C}^1, \boldsymbol{C}^2)}\left( \boldsymbol{C}^h \right) ; \boldsymbol{0}, \tilde{S}_\phi(\boldsymbol{C}^1, \boldsymbol{C}^2) \right).
    \label{eq:variationalapproximation}
\end{equation}

\noindent Here, $(C^1, C^2, C^h)$ denotes an input-output triplet from the training data.
The quantities $\tilde{\mu}_\phi(\boldsymbol{C}_1, \boldsymbol{C}_2)$ and $\tilde{S}_\phi(\boldsymbol{C}_1, \boldsymbol{C}_2)$ represent the mean and covariance of the homogenized constitutive coefficients predicted by the VDMN.
We utilize this expression directly in the loss during training.
We observed that employing the Riemannian formulation improves the alignment between the analytic uncertainty estimates and the empirical distributions obtained by sampling the VDMN's variational parameters and propagating then through deterministic DMNs (see Supplementary Note 4 in Supplementary Information).

\subsubsection{Online Prediction}
\label{sec:vdmn_online_mode}

The close agreement between analytic uncertainty predictions and results from direct propagation (see Figure \ref{fig:cs1_linearperformance}, bottom row) confirms that the VDMN serves not only as a variational approximation of the output density, but also as a variational distribution over the ensemble of possible DMN configurations.
Therefore, the online prediction procedure mirrors that of the standard DMN algorithm.
During inference, a finite set of deterministic DMNs is sampled from the trained VDMN by sampling the morphological hyperparameters from the trained hypervariational distributions.
Predictions are then computed using standard online inference on these sampled networks.
While this study focuses on sampling‑based inference, extending the analytic homogenization algorithm (Algorithm \ref{alg:stochastic_homogenization}) to enable closed‑form estimation of uncertainty during online performance remains an important direction for future work.

\subsection{Derivation of Uncertainty Propagation Using Taylor Series}
\label{app:derivations}

The framework and experiments developed in this work rely heavily on the analytic estimation of output distributions that arise from the propagation -- or derivation -- of uncertainty through functional mappings.
Within the VDMN, we first address the estimation of uncertainty in the homogenization process of each core DMN building block, where the setting involves stochastic inputs and a deterministic function.
In the second (inverse uncertainty quantification) application, we extended this analysis to the case of stochastic inputs and a stochastic function, capturing uncertainty not only in the inputs but also in the functional mapping itself.
The formulations derived here generalize both scenarios, beginning with the simpler deterministic case.
The approach builds on and extends the derivations introduced by Venkatraman \textit{et al.} \cite{venkatraman2025matse} and Girard \textit{et al.} \cite{girard2002gaussian}.

Let $f_{ij}(\cdot)$ denote a matrix-valued deterministic function (e.g., the DMN homogenization function that outputs stiffness tensors represented in Voigt matrix notation).
Let $x_i$ represent a stochastic input with known mean, $\bar{x}_i$, and covariance, $S_{ij}$.
In the VDMN, the complete input vector $\boldsymbol{x}$ is formed by concatenating of all the input variables utilized in the homogenization, with each variable vectorized prior to concatenation.
The covariance between inputs is assumed to be zero and deterministic variables are assigned covariance equal to zero.

\subsubsection{Deterministic Functions}

We approximately identify the output distribution, $p(f_{ij})$, by its first and second cumulants, $\bar{f}_{ij} = \mathbb{E}_{p(\boldsymbol{f})}[f_{ij}]$ (Eq.~\eqref{eq:mean_estimate}) and $S^{f}_{ijkl} = \mathbb{E}_{p(\boldsymbol{f})}[(f_{ij}-\bar{f}_{ij})(f_{kl}-\bar{f}_{kl})]$ (Eq.~\eqref{eq:variance_estimate}), respectively.
Leveraging the Law of the Unconscious Statistician (LOTUS), these expectations over unknown densities can be reconstituted as expectations over the known inputs densities, $\bar{f}_{ij}(\boldsymbol{x}) = \mathbb{E}_{p(\boldsymbol{x})}[f_{ij}(\boldsymbol{x})]$ and $S^{f}_{ijkl} = \textrm{COV}_{p(x)}[\boldsymbol{f}]_{ijkl} = \mathbb{E}_{p(\boldsymbol{x})}[(f_{ij}(\boldsymbol{x})-\bar{f}_{ij}(\boldsymbol{x}))(f_{kl}(\boldsymbol{x})-\bar{f}_{kl}(\boldsymbol{x}))]$, respectively.
We derive expressions for these quantities using Taylor series analyses.
Einstein summation is used throughout.

\begin{equation}
    f_{ij}(\boldsymbol{x}) = f_{ij}(\bar{\boldsymbol{x}}) + 
    \frac{\partial f_{ij}}{\partial x_a}\bigr|_{\boldsymbol{x} = \bar{\boldsymbol{x}}} (\boldsymbol{x} - \bar{\boldsymbol{x}})_a + 
    \frac{1}{2!} \frac{\partial^2 f_{ij}}{\partial x_a \partial x_b}\bigr|_{\boldsymbol{x} = \bar{\boldsymbol{x}}} (\boldsymbol{x} - \bar{\boldsymbol{x}})_a (\boldsymbol{x} - \bar{\boldsymbol{x}})_b + \mathcal{O}(3).\label{eq:TaylorS}
\end{equation}

\noindent \textbf{\textit{Estimating the Mean}}

The following expressions rely on the linearity of the estimation operator and the fact that the partial derivatives are constant with respect to the input, $\boldsymbol{x}$. 

\begin{equation}
\begin{split}
    \mathbb{E}_{p(\boldsymbol{x})}[f_{ij}(\boldsymbol{x})] &\approx \mathbb{E} \left[ f_{ij}(\bar{\boldsymbol{x}}) \right] + 
    \frac{\partial f_{ij}}{\partial x_a}\bigr|_{\boldsymbol{x} = \bar{\boldsymbol{x}}} \mathbb{E} \left[ (\boldsymbol{x} - \bar{\boldsymbol{x}})_a \right] + 
    \frac{1}{2!} \frac{\partial^2 f_{ij}}{\partial x_a \partial x_b}\bigr|_{\boldsymbol{x} = \bar{\boldsymbol{x}}} \mathbb{E} \left[ (\boldsymbol{x} - \bar{\boldsymbol{x}})_a (\boldsymbol{x} - \bar{\boldsymbol{x}})_b \right] \\
    &\approx f_{ij}(\bar{\boldsymbol{x}}) + 
    \frac{1}{2!} \frac{\partial^2 f_{ij}}{\partial x_a \partial x_b}\bigr|_{\boldsymbol{x} = \bar{\boldsymbol{x}}} S_{ab}. \\
\end{split}
\label{eq:methods_mean_derivation}
\end{equation}

\noindent \textbf{\textit{Estimating the Covariance}}

The follow expressions utilize a second order Taylor series of $f_{ij}(\boldsymbol{x})$ and a first order approximation of the expectation $\mathbb{E} [f_{ij}(\boldsymbol{x})]$.

\begin{align}
    f_{ij}(\boldsymbol{x}) - \mathbb{E} [f_{ij}(\boldsymbol{x})] &\approx f_{ij}(\bar{\boldsymbol{x}}) + 
    \frac{\partial f_{ij}}{\partial x_a}\bigr|_{\boldsymbol{x} = \bar{\boldsymbol{x}}} (\boldsymbol{x} - \bar{\boldsymbol{x}})_a - f_{ij}(\bar{\boldsymbol{x}}) \\
    &\approx \frac{\partial f_{ij}}{\partial x_a}\bigr|_{\boldsymbol{x} = \bar{\boldsymbol{x}}} (\boldsymbol{x} - \bar{\boldsymbol{x}})_a.\label{eq:TaylorSCOV}
\end{align}

\noindent We can use this to estimate the covariance.

\begin{equation}
    \begin{split}
    \textrm{COV}[\boldsymbol{f}(\boldsymbol{x})]_{ijkl} &= \mathbb{E} [(\boldsymbol{f}(\boldsymbol{x}) - \mathbb{E} [\boldsymbol{f}(\boldsymbol{x})])_{ij} (\boldsymbol{f}(\boldsymbol{x}) - \mathbb{E} [\boldsymbol{f}(\boldsymbol{x})])_{kl}] \\
    &\approx \mathbb{E}  \left[ \frac{\partial f_{ij}}{\partial x_a}\bigr|_{\boldsymbol{x} = \bar{\boldsymbol{x}}} (\boldsymbol{x} - \bar{\boldsymbol{x}})_a \frac{\partial f_{kl}}{\partial x_b}\bigr|_{\boldsymbol{x} = \bar{\boldsymbol{x}}} (\boldsymbol{x} - \bar{\boldsymbol{x}})_b \right] \\
    &\approx \frac{\partial f_{ij}}{\partial x_a}\bigr|_{\boldsymbol{x} = \bar{\boldsymbol{x}}} \frac{\partial f_{kl}}{\partial x_b}\bigr|_{\boldsymbol{x} = \bar{\boldsymbol{x}}} \mathbb{E} \left[  (\boldsymbol{x} - \bar{\boldsymbol{x}})_a (\boldsymbol{x} - \bar{\boldsymbol{x}})_b \right] \\
    &\approx \frac{\partial f_{ij}}{\partial x_a}\bigr|_{\boldsymbol{x} = \bar{\boldsymbol{x}}} \frac{\partial f_{kl}}{\partial x_b}\bigr|_{\boldsymbol{x} = \bar{\boldsymbol{x}}} S_{ab}.
    \end{split}
    \label{eq:methods_covar_derivation}
\end{equation}

\subsubsection{Stochastic Functions}

When the mapping function $f_{ij}(\cdot)$ is itself a stochastic function, $f_{ij}(\cdot) = p(y_{ij} | \cdot)$, the analysis must incorporate compounding uncertainty arising both from the stochastic inputs and from the intrinsic randomness of the function.
Reusing notation, let $\bar{f}_{ij}(\cdot)$ and $S_{ijkl}^f$ denote the mean and covariance of the stochastic function, $f_{ij}(\cdot)$, respectively.
In this case, the variability originates from the function's inherent stochasticity rather than solely from its inputs.
Mirroring the deterministic case, we utilize the LOTUS and moment approximation to compute the output probability distribution -- $p(C_{ij} = f_{ij}(\cdot))$ -- that results from passing uncertain inputs into a -- now stochastic -- function.
Our treatment closely follows the formulation from Venkatraman \textit{et al.} to handle of stochastic functions \cite{venkatraman2025matse}.

To estimate the output mean, we substitute $f_{ij}(\cdot)$ in Eq. \eqref{eq:methods_mean_derivation} with the mean function $\bar{f}_{ij}(\cdot)$.
Estimating the output covariance requires expanding the analysis to include the additional uncertainty introduced by the stochastic function itself.
Applying the Law of Total Variance, $\textrm{COV}[C_{ij}] = \textrm{COV}_{p(\boldsymbol{x})}[\bar{f}_{ij}(\boldsymbol{x})]] + \mathbb{E}_{p(\boldsymbol{x})}[COV[f_{ij}](\boldsymbol{x})]$, the first term corresponds to the expression already provided in in Eq.~\eqref{eq:methods_covar_derivation} (requiring only that $\bar{f}_{ij}(\cdot)$ replaces $f_{ij}(\cdot)$).
The second term, representing the expected covariance of the stochastic function, can be obtained using the same expansion technique as Eq. \eqref{eq:methods_mean_derivation} since the covariance itself is a deterministic quantity.

\begin{equation}
    S_{ijkl}^C = \frac{\partial \bar{f}_{ij}}{\partial x_a}\bigr|_{\boldsymbol{x} = \bar{\boldsymbol{x}}} \frac{\partial \bar{f}_{kl}}{\partial x_b}\bigr|_{\boldsymbol{x} = \bar{\boldsymbol{x}}} S_{ab} + S_{ijkl}^f + 
    \frac{1}{2!} \frac{\partial^2 S^f_{ijkl}}{\partial x_a \partial x_b}\bigr|_{\boldsymbol{x} = \bar{\boldsymbol{x}}} S_{ab}.
    \label{eq:methods_stochastic_covar_derivation}
\end{equation}

These techniques and expressions are applied throughout this work to consistently handle uncertainty propagation in both deterministic and stochastic mappings.
In practice, we utilized the second order expressions (i.e., the terms including second derivatives) only when we heuristically observed improved performance.
For most applications -- including training the VDMN -- we found that the first order expressions acquired by simply dropping the Hessian terms was sufficient. 

\subsection{Brief Overview of Data Sources}

We used a combination of experimentally collected and simulated datasets in this study.
To train the VDMN, we utilized two different simulated datasets, each generated from a corresponding ensemble of microstructures.
Both ensembles originate from well‑established spinodal decomposition phase‑field simulations~\cite{dingreville2024benchmarking}.
Specifically, microstructures were sampled from the comprehensive dataset of Dingreville \textit{et al.}.
To sample, we select a target microstructure with a desired topography (near circular inclusions or spinodal patterns.
The remaining 29 microstructures completing each ensemble were randomly selected from the nearest fifty microstructures to the target measured by mean squared distance between their 2-point statistics \cite{torquato, fullwoodsurvey}.
Sampling based on n-point statistical similarity systematically introduces controlled morphological variability, effectively emulating the natural microstructural fluctuations encountered during real manufacturing processes~\cite{brough_processstructure_phasefield, fullwoodsurvey, torquato}.
The resulting ensembles were used to construct the VDMN's training, validation, and test sets, each consisting of elastic homogenization simulations with varying orthotropic phase stiffness tensors.
These homogenized responses were computed using an FFT‑based Direct Numerical Solver (DNS)~\cite{shin2023deep}.
For the Validation study, the same FFT solver was further employed to compute nonlinear mechanical responses under distinct boundary conditions.
In Application 1 (forward uncertainty quantification on 3D printed tensile specimens), we experimentally validated the VDMN's predictive capabilities by additively manufacturing and mechanically testing printed specimens.
Two types of samples are fabricated:
(1) single‑phase tensile bars composed of two distinct polymers and
(2) two‑phase composite tensile bars combining these polymers in the spinodal architectures used for VDMN training.
The single‑phase specimens were mechanically tested to calibrate the constitutive properties of each phase, while the two‑phase composites were tested in tension and compared directly against the VDMN's predicted homogenized stress–strain responses.
Detailed descriptions of the simulation procedures, dataset construction, and experimental protocols are provided in Supplementary Note 1 and Sec. Supplementary Note 5 in Supplementary Information.

\bmhead{Acknowledgements}

The authors would like to thank Lane Schultz, Daniel Vizoso, David Montes de Oca Zapian from Sandia National Laboratories and Aaron Tallman from Flroida International University for their insightful discussions throughout the completion of this work.

\section*{Funding Declaration}
DS acknowledges support from the National Research Foundation of Korea(NRF) grant funded by the Korea government(MSIT) (RS-2025-24534529).
AER, ATL, RD acknowledge support from the Advanced Scientific Computing program at Sandia National Laboratories.
SI and BLB acknowledge support from the Aging \$ Lifetime program at Sandia National Laboratories.
RAL acknowledges support from the Advanced Engineering Materials program at Los Alamos National Laboratory.
The DMN capability and computational resources are supported in part by the Center for Integrated Nanotechnologies, an Office of Science user facility operated for the U.S. Department of Energy.
This article has been authored by an employee of National Technology \& Engineering Solutions of Sandia, LLC under Contract No. DE-NA0003525 with the U.S. Department of Energy (DOE).
The employee owns all right, title, and interest in and to the article and is solely responsible for its contents. The United States Government retains and the publisher, by accepting the article for publication, acknowledges that the United States Government retains a non-exclusive, paid-up, irrevocable, world-wide license to publish or reproduce the published form of this article or allow others to do so, for United States Government purposes.
The DOE will provide public access to these results of federally sponsored research in accordance with the DOE Public Access Plan https://www.energy.gov/downloads/doe-public-access-plan.

\section*{Competing Interest}

The authors declare no competing interests.

\section*{Author Contributions}
A.E.R., D.S., and R.D. conceived the idea and designed the research methods.
A.E.R. implemented the VDM model.
S.B.I. and B.L.B collected the experimental data.
A.E.R and A.T.L. generated the training data.
A.E.R. and R.D. analyzed the results and validated the results.
R.D., R.L., and B.L.B. supervised and guided the research project.
R.D., R.L., and B.L.B. secured funding for the research project.
A.E.R., S.I., and R.D. prepared the initial draft.
All authors reviewed the manuscript.

\begin{appendices}

\end{appendices}

\backmatter

\bmhead{{\Large Supplementary Information}}

\section*{Supplementary Note 1: Norton elastoviscoplasticity constitutive model.}
\addcontentsline{toc}{section}{Supplementary Note 1: Norton elastoviscoplasticity constitutive model.}
This Supplementary Note provides a summary of the Norton elastoviscoplastic constitutive model utilized in the main paper.
\\
We utilized Norton elastoviscoplasticity \cite{lebensohn2013modeling} as the constitutive law for the individual phases in the online -- i.e., nonlinear mechanical predictions -- presented in this paper. The Norton constitutive law is defined by the following coupled equations. 

\begin{equation}
\begin{split}
    \dot{\boldsymbol{\varepsilon}}^e = \dot{\boldsymbol{\varepsilon}} - \dot{\boldsymbol{\varepsilon}}^p = \dot{\boldsymbol{\varepsilon}} - \frac{3}{2}\frac{1}{\zeta} \left( \frac{q}{\zeta} \right)^{N-1} \boldsymbol{s} \\
\end{split}
\end{equation}

\noindent Here, $\dot{\boldsymbol{\varepsilon}}$ is the applied total strain rate at a material point,
$\boldsymbol{s}$ is the deviatoric stress: $\boldsymbol{s} = \boldsymbol{\sigma} - (1/3)\mathrm{tr}(\boldsymbol{\sigma})I_{3\times3}$,
$q$ is proportional to the magnitude of the deviatoric stress: $q=\sqrt{3/2} ||\boldsymbol{s}||$,
$\zeta$ is the reference stress given by $\zeta = \sigma_y + K_p \alpha + (\sigma_y^\mathrm{max} - \sigma_y) [1 - \exp(-\delta\alpha)]$, where 
$\dot{\alpha} = \left(\frac{q}{\zeta} \right)^N$ is an internal variable that mediates hardening in the plastic regime.
Isotropic elasticity is used to relate the elastic strain to the stress.
In total, seven parameters parameterize the isotropic linear elasticity and Norton elastoviscoplasticity models used in this paper:
\begin{enumerate}
    \item Young's modulus: $E$
    \item Poisson's ratio: $\nu$
    \item Initial yield stress: $\sigma_y$
    \item Maximum yield stress: $\sigma_y^\mathrm{max}$
    \item Saturation constant: $\delta$
    \item Linear hardening: $K_p$
    \item Power-law exponent: $N$
\end{enumerate}
The constitutive law was implemented using a implicit forward Euler solver.

\section*{Supplementary Note 2: Graphical calibration tests for uncertainty quantification}
\addcontentsline{toc}{section}{Supplementary Note 2: Graphical calibration tests for uncertainty quantification}
This Supplementary Note introduces S\"ailynoja \textit{et al.}'s \cite{sailynoja2022graphical} graphical calibration test: a graphical statistical test utilized in Figure 2 of the main paper to quantitatively assess whether the VDMN's uncertainty predictions are well calibrated.
It contains a statistical confidence interval test and, thus, is a direct statistical measure of uncertainty performance.
\\
We utilized statistical calibration tests to quantitatively validate that the uncertainties predicted by the VDMN in offline mode were consistent with the uncertainty present in the test dataset itself.
We restricted our analysis to the predicted marginal distributions in order to utilize standard calibration tests. 

To test uncertainty calibration on the held out test dataset, we utilized the mean and covariance predicted for each testing point to whiten the testing point, $z^l_{ij} = (C^l_{ij} - \tilde{\mu}_{ij}^l) / \tilde{S}^l_{ijij}$.
Here, $l$ is the sample index and $i,j$ are the marginal indexes, $\tilde{\mu}$ and $\tilde{S}$ are the VDMN predicted means and covariances.
The resulting z-scores will be unit Gaussian distributed if and only if the model is well calibrated and the uncertainty in the data is well approximated by a multivariate Gaussian distribution, \cite{sailynoja2022graphical, gruich2023clarifying}.
We utilized S\"ailynoja \textit{et al.}'s graphical calibration test to assess whether our model produced unit Gaussian z-scores on the held out testing dataset \cite{sailynoja2022graphical}, Figure 2 of the main paper.
In practice, this is performed by comparing the z-scores' rank statistics -- a statistic that reports the percent of the dataset that is less than the specific test value, Eq.~\eqref{eq:app_rankstatistics} -- against the empirical cumulative distribution function computed at the rank statistic, Eq.~\eqref{eq:app_empiricaldistributionfunction}, of the probability integral transform of the z-score using a unit Gaussian probability distribution function as reference, Eq.~\eqref{eq:app_probabilityintegraltransform}.
A perfect equivalence between these values indicates perfect calibration. 

\begin{equation}
    \tilde{r}^l = \frac{1}{N} \sum_{j=1}^N \mathbb{I}(z^j \le z^l),
    \label{eq:app_rankstatistics}
\end{equation}

\begin{equation}
    u^l = \int_{-\infty}^{z^l} p(x) dx,
    \label{eq:app_probabilityintegraltransform}
\end{equation}

\begin{equation}
    F(\tilde{r}^l) = \frac{1}{N} \sum_{j=1}^N \mathbb{I}(u^j \le \tilde{r}^l).
    \label{eq:app_empiricaldistributionfunction}
\end{equation}

\noindent Here, $N$ is the size of the testing dataset, $\mathbb{I}(\cdot)$ is the indicator function, and $p(\cdot)$ is the probability distribution function for the unit normal. 

We utilized the simulation based confidence interval method described by S\"ailynoja \textit{et al.} to define confidence intervals.
This provides a direct graphical confidence test assuring that any deviation in the equivalence between the two computed statistics is not statistically significant.
In this paper, we reported the $95\%$ confidence interval.
Beyond providing a quick and interpretable graphical equivalence test, this method is especially useful because it facilitates identification of the type of miscalibration present (if any is present).
For example, as discussed by S\"ailynoja \textit{et al.}, consistent errors in the mean or variance lead to easily identifiable deformation to the plot -- e.g., end-to-end bowing and s-curve structure for mean and variance errors, respectively.
%

\section*{Supplementary Note 3: VDMN elastic homogenization datasets}
\addcontentsline{toc}{section}{Supplementary Note 3: VDMN elastic homogenization datasets}
\newcounter{mysectnum}
\setcounter{mysectnum}{3}
\renewcommand{\thesubsection}{S\arabic{mysectnum}.\arabic{subsection}}
\setcounter{subsection}{0}
This Supplementary Note outlines the methods used to build the two different datasets utilized in the two application case studies in the main paper. This note includes descriptions of
(1) how the two microstructure ensembles are curated in order to mimic the variability that is observed in real applications,
(2) how elastic stiffness coefficients are selected for building the training, validation, and testing datasets, and
(3) how the microstructures and selected coefficients are combined to construct the training datasets.
Also provided in this note is a description of the final training strategy utilized for the VDMN as well as descriptions of hyperparameter optimizations.
\\
The VDMN was trained using an elastic homogenization dataset similar to previous DMN implementations \cite{shin2023deep, noels2022micromechanics}.
We utilized the 2D method described in Shin \textit{et al.}~\cite{shin2023deep} to generate a dataset of input elastic constitutive parameters for the two phases. 

\begin{equation}
\begin{split}
    C^{11}_\alpha = 1\\
    ~\ln(C^{11}_\beta / C^{11}_\alpha)\in [-1, 1]\\ ~\ln(C^{22}_\alpha / C^{11}_\alpha)\in [-1, 1];~\ln(C^{22}_\beta / C^{11}_\beta)\in [-1, 1]\\
    ~C^{12}_\alpha / \sqrt{C^{11}_\alpha C^{22}_\alpha}\in [0, 0.9]\\
    ~C^{12}_\beta / \sqrt{C^{11}_\beta C^{22}_\beta}\in [0, 0.9]\\
    ~\ln \left( C^{33}_\alpha / \sqrt{C^{11}_\alpha C^{22}_\alpha} \right)\in [-1, 1]\\
    ~\ln \left( C^{33}_\beta / \sqrt{C^{11}_\beta C^{22}_\beta} \right) \in [-1, 1].
\end{split}
\end{equation}

\noindent Values were sampled from the defined uniform sets using a Latin Hyper Cube (LHS) space filling design.
Standard elastic homogenization simulations were performed using the FFT-based method described in Shin \textit{et al.}\cite{shin2023deep} in order to calculate the homogenized stiffness coefficients that are used as targets during training.
The major deviation from standard DMN dataset construction is that we varied the microstructure that is used to run simulations.
In both of the case studies (see Section 2.2 in the main paper), an ensemble of $30$ sample microstructures represent the system's microstructural variability.
To generate the dataset for each presented application, one of the $30$ microstructure instances were randomly assigned to each of the $1655$ generated sets of input parameters for use in the homogenization simulations.
Notably, this strategy means that each microstructure was only assigned to approximately $55$ input constitutive parameter conditions.
The $1655$ input-output triplets were split $0.7:0.15:0.15$ into train, validation, and test sets after the dataset was generated.
The validation dataset was used for hyperparameter optimization.
The test dataset was held out until the final analysis presented in this paper was performed.

\subsection{Microstructures}
\label{sec:microstructure_datasets}

In both case studies (see Section 2.2 in the main paper), we sourced $30$ microstructures from Dingreville \textit{et al.}'s spinodal decomposition dataset \cite{dingreville2024benchmarking}.
The microstructures were selected by identifying a target microstructure -- selected to display desired complexity -- and appending a random selection of $29$ of the $50$ closest microstructures.
Proximity was measured using an $L_2$ distance between their 2-point statistics.
All microstructures were selected from a single time step, $t=3$.
The $30$ microstructures were visually inspected to ensure that no two were too similar.
This approach of selecting statistically similar microstructures mimics the variability that is expected to be caused by, for example, variability in microstructure processing \cite{brough_processstructure_phasefield, fullwoodsurvey, torquato}.
The microstructures are subsequently segmented and resized for simulation purposes.
In the `Validation of VDMN Predictive Performance on Synthetic
Microstructure Ensembles' study (see Sec.~2.1 in the main paper) and `Inverse Uncertainty Quantification' case study (see Sec.~2.2.2 in the main paper), we target microstructures displaying near circular, droplet like inclusions -- similar to standard microstructures studied in previous DMN efforts.
In the `Forward Uncertainty Quantification on 3D Printed Samples' case study (see Sec.~2.2.1 in the main paper), we target spinodal-type microstructures. Previous research has noted that DMN's struggle modeling spinodal type microstructures \cite{shin2023deep}. 

\subsection{Training}
\label{app:training}

The VDMN training strategy was adapted from the deterministic strategy presented by Shin \textit{et al.} \cite{shin2023deep}.
The network takes as input the stiffness matrices of the two constituent phases and is trained to minimize Eq. (1) in the main paper in order to accurately approximate the distribution of the output homogenized stiffness matrices.
In both cases, the dataset was split $70\%$, $15\%$, and $15\%$ into train, validation, and test, respectively
Unless otherwise noted, networks were trained for $4,000$ epochs using the Adam optimizer \cite{kingma2014adam} with AMSGrad \cite{reddi2019convergence}.
No weight decay is used. We used the standard DMN `sine' learning rate scheduler \cite{liu2019deep}.
We also tried a standard cosine annealing scheduler with warm initialization \cite{loshchilov2016sgdr} but observed more inconsistent performance.
We utilized a standard $1,000$ penalty on the model weight sum to conserve the volume fraction (see Shin \textit{et al.} \cite{shin2023deep} for instance).
In addition, after training, we rescaled the learned weights to precisely enforce that they sum to one. We note that this type of rescaling has no effect on the offline predictions because the DMN performs pair weight normalization to compute the volume fractions before homogenization.
However, it does cause greater agreement between the DMN's offline and online modes (i.e., homogenized elastic moduli computed from predicted stress-strain curves will be in agreement with those predicted directly in offline mode). 

We performed a grid-based hyperparameter optimization over the batch size (128, 256, 512) and learning rate (0.01, 0.001, 0.0001) for all VDMN construction strategies considered. Hyperparameter optimization was performed for a 3-layer-deep DMN.
Optimal hyperparameters were selected based on performance on the validation dataset. For first order networks (see Supplementary Note 4), we identify an optimal learning rate of $0.01$ and batch size equal to $256$.
For second-order networks, the optimal batch size was found to reduce to $128$.


\section*{Supplementary Note 4: VDMN ablation study}
\addcontentsline{toc}{section}{Supplementary Note 4: VDMN ablation study}
\setcounter{mysectnum}{4}
\renewcommand{\thesubsection}{S\arabic{mysectnum}.\arabic{subsection}}
\setcounter{subsection}{0}
This Supplementary Note provides a description of the ablation study analyzing the sensitivity of the framework to specific elements (the order of the Taylor series used for propagation, the use of Riemannian statistics in the loss, and the depth of the VDMN network).
It also introduces a performance comparison between the energy-based methods and the proposed framework.
The framework proposed in the main body of the paper outperforms the alternative energy-based method, but is harder to implement.
We include the alternative method in this note because it relates to recent developments in hypernetworks and, as a result, we believe it might prove useful in future work \cite{chan2024hyperdiffusion}.
\\
We present three ablation studies to inspect the importance of various elements of the proposed VDMN framework.
For all three, we utilized the dataset from Sec.~2.1 in the main paper. 

\begin{figure}[!ht]
    \centering
    \includegraphics[width=1.0\linewidth]{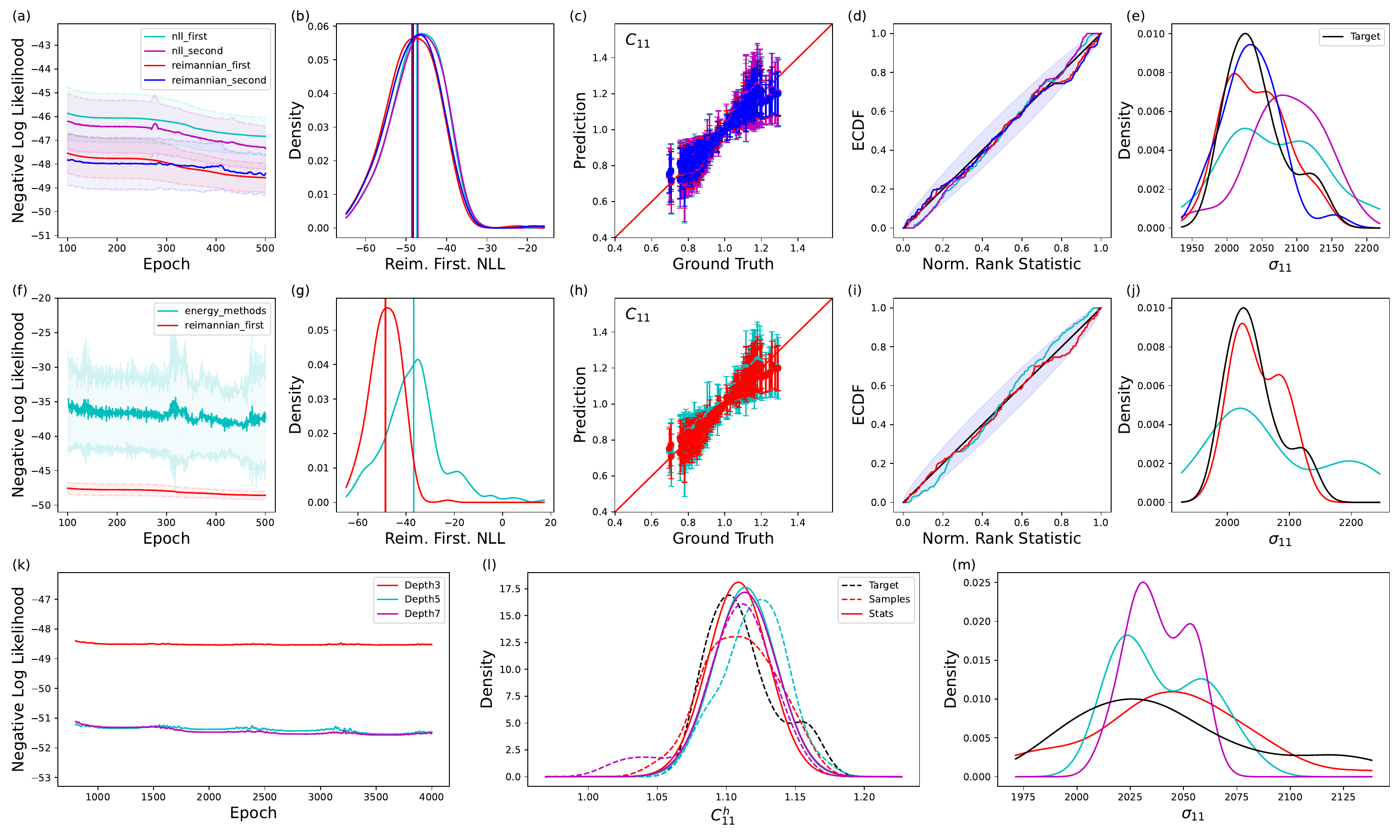}
    \caption{\textbf{Ablation Studies of the Variational Deep Material Network Framework.}
    {\bf Top Row}: Ablation study comparing performance of a Depth 3 VDMN trained using first and second order Taylor series for homogenization as well as using Riemannian and standard Euclidean statistics for distribution approximation. The first order Riemannian variant outperforms the other methods while remaining cheap to train.
    {\bf Middle Row}: Ablation study comparing the performance of a 3-layer-deep First Order Riemannian VDMN against a 3-layer-deep based VDMN. The first order Riemannian VDMN outperforms the energy method and achieves significantly more consistent training.
    {\bf Bottom Row}: Ablation study comparing performance of the First-Order Riemannian VDMN with increasing depth. Color schemes are consistent within each row. In Top and Middle Rows, inference performance is averaged over $15$ unique training runs. In the Bottom Row, a single training run is used. The difference illustrates the remaining variability in the VDMN's performance across training runs and the benefit of ensemble inference when possible.}
    \label{fig:app_ablation_overview}
\end{figure}

\subsection{Framework Ablation}

First, we systematically tested the impact of the order of the Taylor series expansion utilized in the homogenization equations and the type of statistics used to estimate the conditional predicted distribution.
Specifically, we compared the performance of VDMNs trained with both first- and second-order Taylor series constructions (in the mean), using Riemannian statistics and Euclidean statistics.
For simplicity, we restricted ourselves to a 3-layer-deep VDMN and early stop training at $500$ epochs.
For each of the four constructions, we repeated training for $15$ random seeds to study variability in the training.
Supplementary Figures \ref{fig:app_ablation_overview}(a-e) contrast the four ablation settings considered.
The Riemannian variants achieve a better -- i.e., more likely -- approximation of the held out testing data than the standard Euclidean methods; they achieve lower average loss across fifteen repeated training runs, see Supplementary Figure \ref{fig:app_ablation_overview}(a).
However, it is worth noting that the decrease is not immensely significant when compared against the range of loss observed across the dataset, Supplementary Figure \ref{fig:app_ablation_overview}(b).
The primary improvement induced by the Riemannian method is that it is able to identify tighter bounds on the uncertainty than the Euclidean variants. This is observable in the linear (offline setting), see Supplementary Figure \ref{fig:app_ablation_overview}(c), although calibration under any training regime is statistically plausible, see Supplementary Figure \ref{fig:app_ablation_overview}(d).
However, the benefit is more clear in the nonlinear setting where the distributions predicted by the Riemannian networks are in strong agreement, Supplementary Figure \ref{fig:app_ablation_overview}(e).

In contrast, increasing the degree of the Taylor series approximation leads to no observable improvement in the VDMNs performance.
This is immensely beneficial because using higher order Taylor series increases the training time by roughly an order of magnitude due to the expensive Hessian computations.
Additionally, the higher order Taylor series lead to instability in the Riemannian methods.
This is because the second-order Taylor series is not guaranteed to predict a mean stiffness matrix contained in the set of PSD matrices.
When predictions are outside the set, the logarithmic map becomes undefined and causes the training to fail.
We observed that this occurs on roughly $50\%$ of random training runs, but almost always occurs within the first $5$ epochs.
The Euclidean method is unaffected by such instabilities regardless of the order of the Taylor series.
Fortunately, first-order methods appear sufficient and the Riemannian method can be used without concern. 

\subsection{Depth Ablation}

Increasing the depth of a DMN increases its expressibility and has been demonstrated to lead to improved performance \cite{shin2023deep}.
The VDMN is no exception.
Supplementary Figures \ref{fig:app_ablation_overview}(k,l,m) illustrate performance trends with increasing depth.
The VDMN displays improved performance with increasing depth, see Supplementary Figure \ref{fig:app_ablation_overview}(k).
For this example, the loss improvement saturates almost completely for a 5-layer-deep network.
Salient characteristics of the VDMN model remain stable with this scaling.
The VDMN's sampling and analytic mode remain in agreement with increasing depth, see Supplementary Figure \ref{fig:app_ablation_overview}(l).
Further, extrapolation performance in the nonlinear regime remains stable, see Supplementary Figure \ref{fig:app_ablation_overview}(m).
We note that this comparison is performed with just a single run, thus the disagreement between Supplementary Figures \ref{fig:app_ablation_overview}(e) and (j) and Supplementary Figure \ref{fig:app_ablation_overview}(m).
In general, we found that training and running inference using an ensemble of VDMNs consistently improved performance.
Notably, this ensemble sampling has no effect on the computational cost for sample-based inference.

\subsection{Comparison against Energy-based Methods}

Beyond the final method proposed here, we developed alternative approaches that also displayed promise for training the VDMN's hypervariational parameters.
The most promising discarded variant was based on energy methods (i.e., Maximum Mean Discrepency (MMD) methods \cite{cui2020calibrated, li2015generative, borgwardt2006integrating}).
This is a purely sample based strategy that does not require the analytic distribution techniques described in this paper.
Similar ideas have also been used successfully in other probabilistic methods in scientific machine learning \cite{generale2023bayesian}.
Although this makes writing the training code for this technique easier, it also limits the VDMN's utility by removing the ability to directly estimate distributions.
This, for example, would make the second presented application -- inverse uncertainty quantification -- significantly more computationally challenging to implement. 

An energy-based training method is constructed by observing that training the VDMN to match the conditional distribution $P(C^h|C^\alpha, C^\beta)$ on a sample by sample basis results in an equivalent VDMN as training to match the joint distribution $P(C^h, C^\alpha, C^\beta)=P(C^h|C^\alpha, C^\beta)P(C^\alpha, C^\beta)$ on a batch by batch basis.
Notably, here, the prior distribution is defined implicitly by the training data itself.
Under this training strategy, a single prediction $C^h$ is randomly sampled from the VDMN for each input $\{C^\alpha, C^\beta\}$ -- this is performed by randomly sampling the DMN parameters from the hypervariational distributions and then running standard DMN inference.
Subsequently, the MMD distance is computed between the set of sampled $\{C^{h,b}\}_{b \in B}$ values and the set of ground truth values $\{C^{*,h,b}\}_{b \in B}$.
The inputs, $\{C^{\alpha, b}, C^{\beta, b}\}_{b \in B}$, are not included in the calculation because they are guaranteed to be in agreement.
It is extremely important to emphasize that the MMD distance is not averaged across samples in the batch, but rather the MMD distance computation includes all points in the batch \textit{concurrently}. 

We trained $15$ Depth-3 VDMN using the proposed energy based method.
Like before, we performed hyperparameter optimization over the batch size, learning rate, and MMD distance parameters.
We identified the following optimal values:
learning rate: $0.001$,
batch size: $128$, and
MMD kernel parameters: $[0.01, 0.05, 0.5, 5.0]$. 

The proposed VDMN training framework out-performs the energy-based method, see Supplementary Figure \ref{fig:app_ablation_overview} middle row.
It achieves a significantly lower negative log likelihood.
Additionally, the energy methods display greater training variability, indicated by the larger variance in the loss curves.
These errors lead to over estimation of the predicted uncertainty bounds, Supplementary Figure \ref{fig:app_ablation_overview}(h,i).
Further, the proposed VDMN displays more accurate extrapolation to nonlinear predictions, see Supplementary Figure \ref{fig:app_ablation_overview}(j).

In total, the proposed VDMN method provides improved performance as well as analytic uncertainty estimates valuable for downstream tasks.
Furthermore, training cost is nearly identical between the two methods.
However, we include the energy based method because we found that it is significantly easier to adapt existing DMN implementations to the VDMN format using this strategy. Therefore, we believe that this alternative training paradigm retains value as an exploratory strategy.
Additionally, its relation to recent research in hypernetworks may provide useful in future total uncertainty quantification efforts \cite{chan2024hyperdiffusion}.

\section*{Supplementary Note 5: Additional information on forward uncertainty quantification case study}
\addcontentsline{toc}{section}{Supplementary Note 5: Additional information on forward uncertainty quantification case study}
\setcounter{mysectnum}{5}
\renewcommand{\thesubsection}{S\arabic{mysectnum}.\arabic{subsection}}
\setcounter{subsection}{0}
This Supplementary Note provides expanded description of the techniques used in Sec.~2.2.1 `Forward Uncertainty Quantification of 3D Printed Samples.'
Specifically, it includes descriptions of the experimental methods, the Bayesian optimization based calibration used to calibrate the Norton elastoviscoplastic model, and adaptations to the DMN’s nonlinear solver to model the mixed boundary conditions present in tensile testing.
The application presented in Sec.~2.2.1 of the main manuscript leverages the VDMN to predict the aleatoric uncertainty in the mechanical response of a two-phase composite printed sample without the need for mechanical testing the heterogeneous components.
Here, the VDMN's predicted uncertainty approximates the aleatoric uncertainty arising due to microstructural variability.
Other sources of uncertainties -- uncertainty in the constituent properties of the individual phases and measurement uncertainty -- were assumed to be negligible; their analysis is left as the subject of the second case study in Sec.~2.2.2 of the main manuscript and Supplementary Note 6.

\begin{figure}[!ht]
    \centering
    \includegraphics[width=1.0\linewidth]{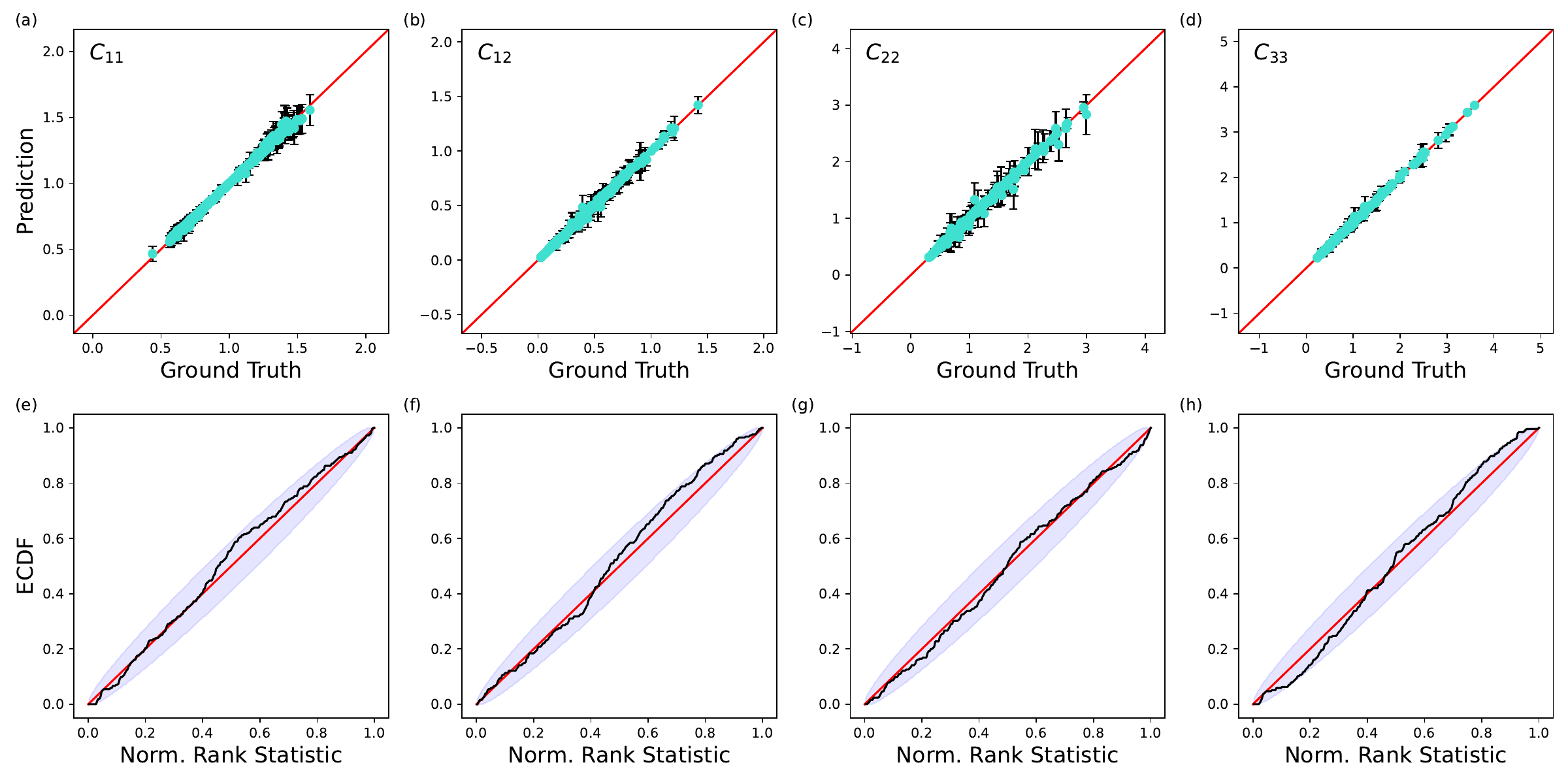}
    \caption{\textbf{Overview of Application 1 -- Spinodal Microstructures -- Offline (i.e., Linear) VDMN Performance.}
    (a)-(d) parity plots comparing the trained VDMN's mean and variance predictions on four homogenized stiffness components against ground truth values computed using FFT-based simulations. The parity line remains consistently within the $3$ standard deviations shown. Compared to the Validation study, in this case study the VDMN identifies significantly tighter bounds on the predicted uncertainties.
    (e)-(h) $95\%$ confidence graphical calibration test \cite{sailynoja2022graphical} of the VDMN's uncertainty predictions, see Supplementary Note 2. The black line quantifies the uncertainty predictions over the test dataset. The red line represents a perfectly calibrated model. The purple envelope is the $95\%$ confidence interval indicating that any deviation from a perfect calibration is likely random. Just as the Validation study (see Sec. 2.1 in the main paper), the VDMN's predictions are well calibrated across all components.}
    \label{fig:si_forwarduq_linearperformance}
\end{figure}

We performed this case study utilizing experimental data collected from mechanical testing of 3D printed samples.
Each sample is a tensile test sample comprised of a variable repeating unit cell.
We studied a two-phase composite microstructure comprised of two polymer phases, see Sec. \ref{sec:si_sam_experimental_setup}.
To emulate the microstructure variability that can arise due to control limits in high volume manufacturing, we imposed an ensemble of $30$ statistically similar composite unit cells derived from spinodal decomposition phase-field simulations.
The unit cells were selected by their proximity to a desired microstructure unit cell, where proximity was measured by Euclidean distance between their 2-point statistics, see Sec. \ref{sec:microstructure_datasets}.
Examples of the spinodal microstructure unit cells are depicted in Figure 4 in the main paper and Supplementary Figure \ref{fig:si_tensile_experiment_dimensions}.
We utilized the presented VDMN training strategy to train a VDMN to predict the consequences of the aleatoric variability in this ensemble, see Sec. \ref{app:training}.
We trained the model on $1190$ elastic homogenization simulations calculated via an FFT-based solver.
Supplementary Figure \ref{fig:si_forwarduq_linearperformance} summarizes the offline training performance of the VDMN trained on the spinodal decomposition microstructures.
We observe similar performance trends with those observed in the Validation study (Figure 3 in the main paper).
Finally, we utilized the trained VDMN to predict the expected mechanical response of the composite tensile bars ensemble during experimental tensile testing (i.e., after the bars are printed).
We printed and mechanically tested tensile bar for each polymer phase in order to calibrate the local constitutive models of each phase, see Sec. \ref{sec:si_material_calibration}.
This single phase mechanical testing was the only experimental preparation needed for the VDMN's predictions (we note that in general applications microscopy to quantify the microstructure variability of the composite tensile bars would also be necessary, but is not necessary here because the variability is prescribed to facilitate analysis).
Subsequently, we used the VDMN to simulate and predict the expected distribution of nonlinear responses for the ensemble of composite samples under tensile testing conditions, Sec. \ref{sec:si_tensilesimulations_dmn}.

To validate the VDMN's predictions, we ran uniaxial tensile testing on tensile bars comprised of the selected units cells, see Sec. \ref{sec:si_sam_experimental_setup}.
Specifically, we downselected the $12$ most diverse unit cells for printing.
Here, selection was made by a max-min selection process on their $C_{11}$ homogenized stiffness coefficient \cite{kennard1969maxmin, mak2018maxmin}. 
Briefly, in this process, the candidate microstructure whose minimum distance to the existing dataset is maximized amongst all candidates was subsequently added to the dataset.
The dataset was expanded until a desired size.
Briefly, we note that this downsampling process effectively biased the output distribution towards the uniform compared to the microstructure distribution used for training.
This effect is subtly observable in Figure 4 of the main paper.

\subsection{Experimental Set-Up}
\label{sec:si_sam_experimental_setup}

\begin{figure}[!ht]
    \centering
    \includegraphics[width=1.0\linewidth]{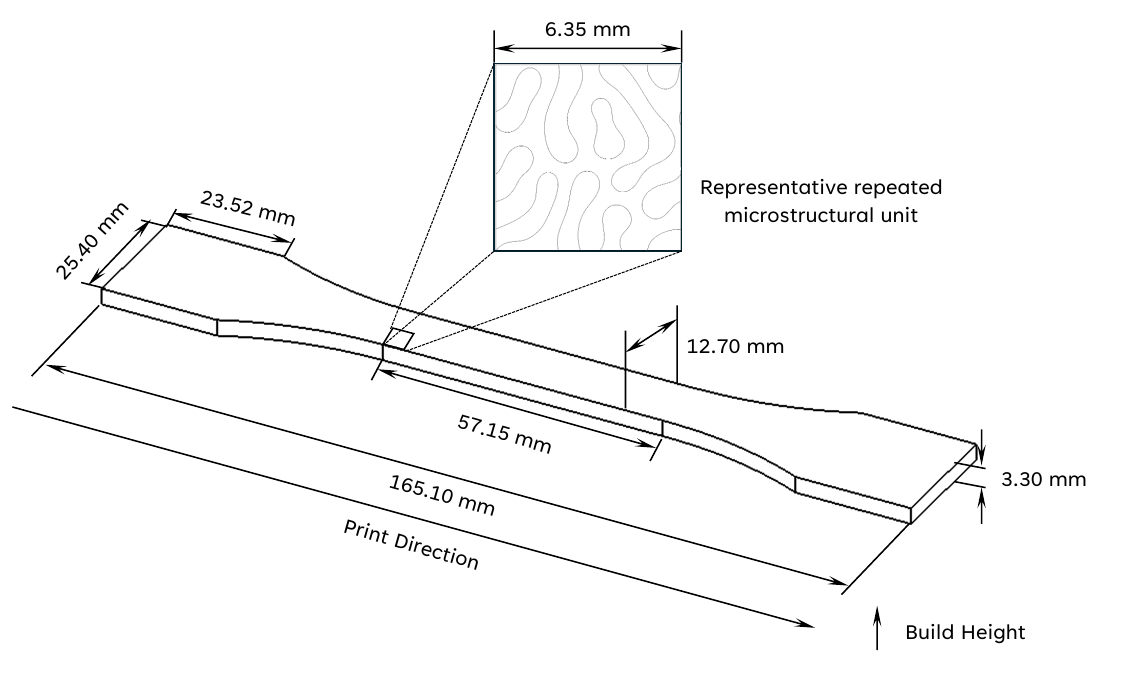}
    \caption{\textbf{Physical Dimensions of Tensile Testing Sample.} Dimensions of the 3D printed tensile test samples compliant with ASTM D638. Three types of samples were printed: (1) single phase (precipitates), (2) single phase (matrix), and (3) two-phase composite. For the two phase composite, the prescribed spinodal unit cell -- shown in the inset -- were repeated periodically throughout the domain.}
    \label{fig:si_tensile_experiment_dimensions}
\end{figure}

Tensile bar dimensions were defined by ASTM D638 with dimensions shown in Supplementary Figure \ref{fig:si_tensile_experiment_dimensions}.
Composite microstructures were produced via spinodal decomposition phase field simulations (see Sec. \ref{sec:microstructure_datasets}) with the phase boundaries manually traced into repeated 6.35 mm square unit cells repeated across the length and width of the tensile bars.
Single phase and multi-phase (composite) tensile bars were printed for mechanical testing.
Both single phase and multi-phase tensile bars were printed in a Stratasys $\mathrm{J826}^{\mathrm{TM}}$ UV-cured material jetting printer, with pure $\mathrm{Vero}^\mathrm{TM}$ cyan forming the particles phase and an approximately equal mixture of $\mathrm{Vero}^\mathrm{TM}$ magenta and $\mathrm{Agilus30}^\mathrm{TM}$ white forming the comparatively less stiff matrix phase (higher volume fraction phase).
The print direction was aligned with the tensile direction, while the build height was oriented parallel to the thickness of the bars.
All bars were power washed with water to remove excess $\mathrm{SUP706B}^\mathrm{TM}$ support material and were allowed to dry for at least 24 hours prior to mechanical testing.

A TestResources $\mathrm{313}^\mathrm{TM}$ servoelectric load frame with an Interface 1110 $\mathrm{AF}^\mathrm{TM}$ 10,000 lbs load cell and MTS $\mathrm{492}^\mathrm{TM}$ mechanical testing software was used for tensile tests on each sample.
The bars were held with a spring-loaded wedge grip preloaded to 20 N connected to a universal joint to maintain alignment.
A constant displacement rate of $0.153$ $mm/s$ (approximately equivalent to a strain rate of $0.003$ $1/s$) was applied until fracture.
Strain was monitored using digital image correlation, following conventional speckling techniques, with live imaging provided by a Teledyne Technologies 12.3 MP monochrome camera and analysis conducted with Correlated Solutions VIC-Gauge 2-DTM software. 

\begin{figure}[!ht]
    \centering
    \includegraphics[width=1.0\linewidth]{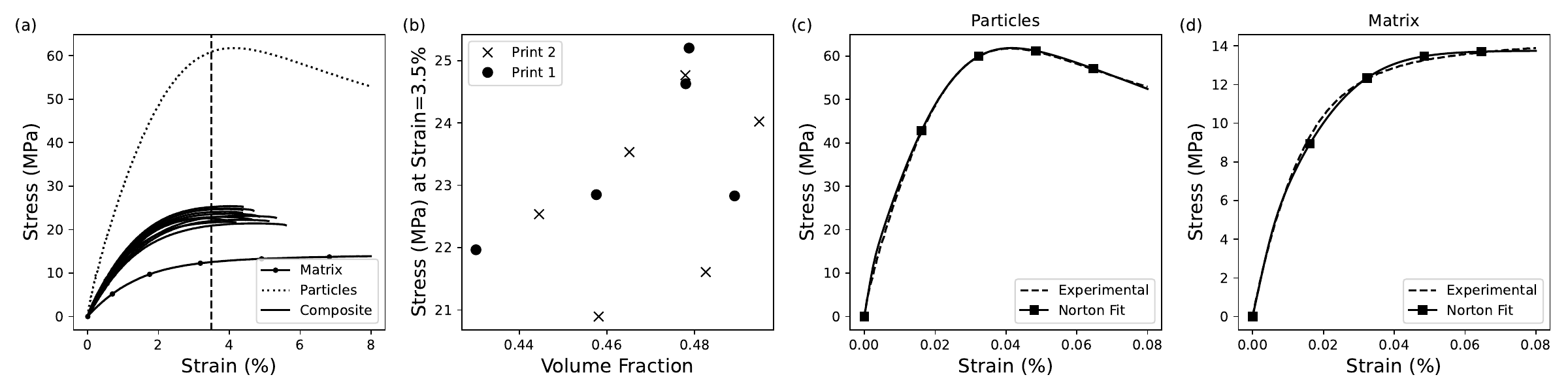}
    \caption{\textbf{Experimental Stress-Strain Measurements and Calibration.}
    (a) Comparison of the experimentally measured stress-strain response for the soft matrix phase (black with dots), the hard particle phase (black dotted), and the ensemble of two-phase composites with varying spinodal unit cell (black line).
    (b) Trend between the measured stress (at $\varepsilon_{xx} = 0.035$) and the particle volume fraction in the composite components. Clear trend indicates that the variability is not just experimental noise. Samples are divided by print to illustrate that the print batch is not a confounding variable in the variability.
    (c) Comparison of the tensile test response between the experimental measurements and the prediction by the calibrated Norton elastoviscoplastic constitutive model for the particle phase (calibrated using Bayesian optimization).
    (d) Comparison of the tensile test response between the experimental measurements and the prediction by the calibrated Norton elastoviscoplastic constitutive model for the matrix phase (calibrated using Bayesian optimization).}
    \label{fig:si_experimentalmeasurementsummary}
\end{figure}

Supplementary Figures \ref{fig:si_experimentalmeasurementsummary}(a,b) summarize the measured stress strain curves for the single phase materials and the composite bars.
Supplementary Figure \ref{fig:si_experimentalmeasurementsummary}(b) illustrates an observable trend between the stress in the nonlinear regime and the volume fraction of the composite unit cell.
The presence of this trend indicates that the measured uncertainty in the ensemble of composite bars does not arise from just experimental measurement error. 

\subsection{Modeling Tensile Testing using the VDMN}
\label{sec:si_tensilesimulations_dmn}

Simulating standard mechanical tensile testing involves solving a homogenization problem under mixed boundary conditions.
Specifically, considering a 2D plane strain approximation, the body experiences displacement control on only a subset of axes: the imposed strain rate along the testing direction, $\dot{\varepsilon}_{xx}=\dot{\varepsilon}$, as well as $\dot{\varepsilon}_{xz}=\dot{\varepsilon}_{yz}=\dot{\varepsilon}_{zz}=0$ to enforce plane strain conditions.
The remaining surfaces are subject to traction free equilibrium conditions, $\sigma_{xy}=\sigma_{yy}=0$.
We utilized a plane strain approximation because the height of the bar (distance along the build height) is large compared to the size of the unit cell (roughly 1:2, respectively), see Supplementary Figure \ref{fig:si_tensile_experiment_dimensions}.
For this reason and because the the 2D unit cell is unchanged along the build height, we argue that the unit cells will reasonably experience plain strain conditions.

The standard DMN derivations are developed to enable completely displacement controlled simulations, i.e., simulations in which the strain rate in each tensor component is prescribed.
In order to simulate tensile testing under mixed conditions using the VDMN, we utilized an exterior Newton-Raphson loop to update the unknown $\dot{\varepsilon}_{xy}$ and $\dot{\varepsilon}_{yy}$ boundary conditions until the traction free, $\sigma_{xy}=\sigma_{yy}=0$, condition on the homogenized stress response is achieved.
A similar method was used by Shin \textit{et al.} \cite{shin2024thermomechanical}.
The unknown conditions are updated using the following expression. 

\begin{equation}
    \begin{bmatrix} \delta \dot{\varepsilon}_{xy} \\ \delta \dot{\varepsilon}_{yy} \end{bmatrix} = -\begin{bmatrix} \frac{\partial \sigma_{xy}}{\partial{\dot{\varepsilon}_{xy}}} & \frac{\partial \sigma_{xy}}{\partial{\dot{\varepsilon}_{yy}}} \\ \frac{\partial \sigma_{yy}}{\partial{\dot{\varepsilon}_{xy}}} & \frac{\partial \sigma_{yy}}{\partial{\dot{\varepsilon}_{yy}}} \end{bmatrix}^{-1} \begin{bmatrix} \sigma_{xy} \\ \sigma_{yy} \end{bmatrix}.
\end{equation}

\noindent The Jacobian matrix was calculated using automatic differentiation via PyTorch's Functorch capability \cite{functorch2021}.
Roughly three exterior Newton-Raphson iterations were needed per global strain step to achieve convergence with $\sigma_{xy}, \sigma_{yy} \le1\times 10^{-12}$. 

\subsection{Calibrating the Individual Phase Constitutive Law}
\label{sec:si_material_calibration}

Seven model parameters for the Norton elasto-viscoplastic constitutive law were calibrated for each single phase material using the single phase stress-strain curves displayed in Supplementary Figure \ref{fig:si_experimentalmeasurementsummary}(a), see Supplementary Note 1.
The calibration was performed using Bayesian optimization via the Scikit-Optimize package \cite{skopt}.
The optimization was performed using $100$ warm up calls (seeded using a sobol process) and $50$ search steps using the `gp-hedge' acquisition function.
Candidate points were selected from a new pool of $10,000$ candidates every iteration. The optimization minimized the normalized absolute error between the stress-strain curve predicted by the Norton elasto-viscoplastic model and the experimental measurement. 

\begin{equation*}
    E = \sum_{i=1}^{100} \frac{|\sigma_{i} - \sigma^{\mathrm{Norton}}_i|}{|\sigma_i|}.
\end{equation*}

\noindent Both stress-strain curves were resampled using linear interpolation onto $100$ equally spaced strain increments to equally weight contribution to the error across the stress-strain curve.

For the soft matrix phase, the final identified parameters were:
Young's modulus: $647.066$ MPa,
Poisson's ratio: $0.49$,
initial yield stress: $45.0$,
maximum yield stress: $90.0$,
saturation constant ($\delta$): $108.677$,
linear hardening ($K_p$): $1.311$,
power-law exponent ($N$): $3$.

For the hard precipitate phase, the final identified parameters were:
Young's modulus: $3749.325$ MPa,
Poisson's ratio: $0.48$,
yield stress: $64.644$,
maximum yield stress: $283.036$,
saturation constant ($\delta$): $109.048$,
linear hardening ($K_p$): $-989.328$,
power-law exponent ($N$): $4$.
Comparison of the predicted stress strain curve against the experimental measurement can be seen in Supplementary Figure \ref{fig:si_experimentalmeasurementsummary}(c,d).
~\clearpage

\section*{Supplementary Note 6: Additional information on inverse uncertainty quantification case study}
\addcontentsline{toc}{section}{Supplementary Note 6: Additional information on inverse uncertainty quantification case study}
\setcounter{mysectnum}{6}
\setcounter{subsection}{0}
\renewcommand{\thesubsection}{S\arabic{mysectnum}.\arabic{subsection}}
This Supplementary Note provides expanded description of the techniques used in Section 2.2.2 `Inverse Uncertainty Quantification.'
Specifically, it includes descriptions of the construction of the total uncertainty model (combining uncertainty in the constitutive laws, microstructure, and experimental measurements) and ablation studies analyzing how which homogenized measurements are performed effects the technique's ability to identify the unknown constitutive law parameter distributions.
\\
This second case study leverages the VDMN to estimate immeasurable sources of uncertainty.
Such sources can occur, for example, at lower length-scales.
In this case study, we utilized the VDMN to estimate the unknown distribution of constitutive parameters associated with each phase in the microstructure.
We performed this estimation in the presence of microstructure and measurement uncertainty.
To demonstrate performance in this situation, we numerically varied the input constituent properties of each phase:
we assumed isotropic elastic coefficients and assigned $E_1 = 1 \pm0.05$ and $E_2 = 2 \pm 0.15$.
The Poisson's ratios were held constant at $\nu_1=0.3$ and $\nu_2=0.19$.
We created a dataset of $30$ synthetic measurements by repeatedly sampling the constitutive parameters and assigning them to the $30$ embedded circle microstructures from the Validation study (Sec. 2.1 in the main paper) and the second case study (Sec. 2.2.2 in the main paper).
Note, we observed similar trends and performance on the spinodal microstructure system from the first case study (Sec. 2.2.1 in the main paper).
We used the FFT-based solver to `measure' homogenized elastic coefficients. 

\subsection{Total Uncertainty Quantification}
\label{app:totaluncertaintymodel_forinverse}

The expected homogenized stiffness distribution -- accounting for measurement uncertainty, microstructural variability, and variability in the elastic constitutive properties of the individual phases -- was estimated by utilizing the pre-trained VDMN in the Uncertainty Propagation with Stochastic Functions framework described in the Methods Section, see Sec. 4.3 of the main paper. 

We utilized the current estimate of the mean and variances of the input constitutive properties as well as Eq. (14) and Eq. (17) in the main paper to estimate the noise free distribution of homogenized stiffnesses (the distribution on compliances -- which are easier to measure experimentally -- can be easily estimated via inverting the predicted stiffnesses and applying Eq. (17) in the main paper to estimate the covariance).
Finally, the effect of the experimental noise was added using a standard Gaussian noise model (i.e., direct addition of the variance times the identity to the predicted noise free covariance).

\subsection{Identifying Valuable Experiments for Inverse Modeling}
\label{app:identifyingvaluableexpforinverse}

We found that taking multiple different homogenized measurements from each sample significantly improved the ability to effectively deconvolve the effect of the measurement noise from the lower length-scale constitutive noise.
This improvement occurs because we can leverage the VDMN's ability to predict the correlation between each measured property.
Because it is often challenging to collect every homogenized coefficient, we collect a subset.
We repeated the optimization process to maximize the log-likelihood of the collected homogenized data and estimate the unknown constitutive distributions using different combinations of measured constitutive properties.
We utilized a simple Nelder-Mead optimizer for all measurement combinations except experiments that contained both $\{ C^h_{11}, C^h_{22} \}$ (here, we used a Conjugate Gradient solver).
Supplementary Table \ref{table:simulation_coefficientvalue} reports the observed results.
Simultaneous measurements of the homogenized tensile and Poisson's response were the most valuable for estimating the distribution of input constituent values.
We observed the same result when repeating the inverse analysis for the microstructure in the first case study: `Forward Uncertainty Quantification on 3D Printed Samples'.

\begin{table}[htbp]  
    \centering
    \begin{tabular}{@{}lcccc@{}}
        \hline
        \midrule
        Components & $\mu_{E_1}: 1.00$ (GPa) & $\log S_{E_1}: -5.99$ (GPa) & $\mu_{E_2}: 2.00$ (GPa) & $\log S_{E_2}: -3.79$ (GPa) \\
        \midrule
        \multicolumn{5}{c}{\textbf{Combined Experiments}} \\ 
        \midrule
        $C^h_{11}$
                                     & $1.07$
                                     & $-5.05$
                                     & $1.53$ 
                                     & $-5.66$ \\
        $C^h_{12}$                        
                                     & $1.09$
                                     & $-5.25$
                                     & $1.52$ 
                                     & $-5.37$ \\
        $C^h_{22}$                        
                                     & $1.07$
                                     & $-5.04$
                                     & $1.53$ 
                                     & $-5.65$ \\
        $C^h_{33}$                        
                                     & $1.04$
                                     & $-5.03$
                                     & $1.67$ 
                                     & $-5.58$ \\
        ${C^h_{11}, C^h_{12}}$                        
                                     & ${0.99}$
                                     & ${-5.50}$
                                     & ${2.01}$ 
                                     & ${-3.79}$ \\
        ${C^h_{12}, C^h_{22}}$                        
                                     & ${0.99}$
                                     & ${-5.51}$
                                     & ${2.02}$ 
                                     & ${-3.71}$ \\
        
        $C^h_{11}, C^h_{33}$                        
                                     & $0.97$
                                     & $-5.63$
                                     & $2.12$ 
                                     & $-7.00$ \\
        
        $C^h_{22}, C^h_{33}$                        
                                     & $0.98$
                                     & $-5.58$
                                     & $2.05$ 
                                     & $-7.00$ \\
        
        $C^h_{12}, C^h_{33}$                        
                                     & $0.99$
                                     & $-5.39$
                                     & $1.99$ 
                                     & $-5.20$ \\
        
        $C^h_{11}, C^h_{12}, C^h_{33}$                        
                                     & $1.00$
                                     & $-5.44$
                                     & $1.94$ 
                                     & $-4.47$ \\
                                     
        $C^h_{11}, C^h_{22}, C^h_{33}$                        
                                     & $1.02$
                                     & $-5.38$
                                     & $1.83$ 
                                     & $-5.01$ \\
        
        $C^h_{11}, C^h_{12}, C^h_{22}, C^h_{33}$                        
                                     & $1.02$
                                     & $-5.35$
                                     & $1.86$ 
                                     & $-5.00$ \\
                                     
        \midrule
        \multicolumn{5}{c}{\textbf{Single Experiments}} \\ 
        \midrule
        $C^h_{11}, C^h_{12}$                        
                                     & $1.10$
                                     & $-5.32$
                                     & $1.47$ 
                                     & $-5.38$ \\
        $C^h_{12}, C^h_{22}$                        
                                     & $1.04$
                                     & $-4.95$
                                     & $1.62$ 
                                     & $-5.79$ \\
        $C^h_{11}, C^h_{33}$                        
                                     & $1.03$
                                     & $-4.89$
                                     & $1.68$ 
                                     & $-5.77$ \\
        $C^h_{22}, C^h_{33}$                        
                                     & $1.03$
                                     & $-4.89$
                                     & $1.68$ 
                                     & $-5.77$ \\
        $C^h_{12}, C^h_{33}$                        
                                     & $1.03$
                                     & $-4.89$
                                     & $1.68$ 
                                     & $-5.77$ \\
        $C^h_{11}, C^h_{12}, C^h_{33}$                        
                                     & $1.00$
                                     & $-4.72$
                                     & $1.78$ 
                                     & $-5.73$ \\
        $C^h_{11}, C^h_{22}, C^h_{33}$                        
                                     & $1.00$
                                     & $-4.72$
                                     & $1.78$ 
                                     & $-5.73$ \\
        $C^h_{11}, C^h_{12}, C^h_{22}, C^h_{33}$                        
                                     & $1.02$
                                     & $-4.60$
                                     & $1.79$ 
                                     & $-6.03$ \\
                                     
        \midrule
        \bottomrule
    \end{tabular}
    \caption{\textbf{Results of Uncertainty Aware Calibration of Uncertain Constitutive Properties with Varying Homogenized Measurements.} This table reports the estimated mean and variance for the Young's Moduli of each constituent phase. The ground truth values are reported at the top. Estimates are made by running the negative loglikelihood optimization described in Sec.~2.2 and Supplementary Note 5 with access to the homogenized experimental measurements enumerated under `Components'. `Combined Experiments' refer to situations where all enumerated `Components' were measured \textit{on each sample}. `Single Experiments' refers to the situation were only one of the enumerated `Components' was measured on each sample. Measurements were performed via simulation. Clearly, multiple measurements per sample improves performance by enabling the optimization to utilize the full joint distributions predicted by the VDMN.}
    \label{table:simulation_coefficientvalue}
\end{table}
~\clearpage
%
%


\bibliography{cas-refs}

@article{shin2023deep,
  title={Deep material network via a quilting strategy: visualization for explainability and recursive training for improved accuracy},
  author={Shin, Dongil and Alberdi, Ryan and Lebensohn, Ricardo A and Dingreville, R{\'e}mi},
  journal={npj Computational Materials},
  volume={9},
  number={1},
  pages={128},
  year={2023},
  publisher={Nature Publishing Group UK London}
}

@article{shin2024thermal,
  title={Deep material network for thermal conductivity problems: Application to woven composites},
  author={Shin, Dongil and Creveling, Peter Jefferson and Roberts, Scott Alan and Dingreville, R{\'e}mi},
  journal={Computer Methods in Applied Mechanics and Engineering},
  volume={431},
  pages={117279},
  year={2024},
  publisher={Elsevier}
}

@article{shin2024thermomechanical,
  title={A deep material network approach for predicting the thermomechanical response of composites},
  author={Shin, Dongil and Alberdi, Ryan and Lebensohn, Ricardo A and Dingreville, R{\'e}mi},
  journal={Composites Part B: Engineering},
  volume={272},
  pages={111177},
  year={2024},
  publisher={Elsevier}
}

@article{liu2019deep,
  title={A deep material network for multiscale topology learning and accelerated nonlinear modeling of heterogeneous materials},
  author={Liu, Zeliang and Wu, CT and Koishi, M391298807188754},
  journal={Computer Methods in Applied Mechanics and Engineering},
  volume={345},
  pages={1138--1168},
  year={2019},
  publisher={Elsevier}
}

@article{wu2025stochastic,
  title={Stochastic deep material networks as efficient surrogates for stochastic homogenisation of non-linear heterogeneous materials},
  author={Wu, Ling and Noels, Ludovic},
  journal={Computer Methods in Applied Mechanics and Engineering},
  volume={441},
  pages={117994},
  year={2025},
  publisher={Elsevier}
}

@article{gajek2020micromechanics,
  title={On the micromechanics of deep material networks},
  author={Gajek, Sebastian and Schneider, Matti and B{\"o}hlke, Thomas},
  journal={Journal of the Mechanics and Physics of Solids},
  volume={142},
  pages={103984},
  year={2020},
  publisher={Elsevier}
}

@article{noels2022micromechanics,
  title={Micromechanics-based material networks revisited from the interaction viewpoint; robust and efficient implementation for multi-phase composites},
  author={Nguyen, Van Dung and Noels, Ludovic},
  journal={European Journal of Mechanics-A/Solids},
  volume={91},
  pages={104384},
  year={2022},
  publisher={Elsevier}
}

@article{wei2025orientation,
  title={Orientation-aware interaction-based deep material network in polycrystalline materials modeling},
  author={Wei, Ting-Ju and Su, Tung-Huan and Chen, Chuin-Shan},
  journal={arXiv preprint arXiv:2502.02457},
  year={2025}
}

@article{wei2024foundation,
  title={Foundation Model for Composite Materials and Microstructural Analysis},
  author={Wei, Ting-Ju and others},
  journal={arXiv preprint arXiv:2411.06565},
  year={2024}
}

@article{huang2022microstructurefibervariation,
  title={Microstructure-guided deep material network for rapid nonlinear material modeling and uncertainty quantification},
  author={Huang, Tianyu and Liu, Zeliang and Wu, CT and Chen, Wei},
  journal={Computer Methods in Applied Mechanics and Engineering},
  volume={398},
  pages={115197},
  year={2022},
  publisher={Elsevier}
}

@article{dey2022training,
  title={Training deep material networks to reproduce creep loading of short fiber-reinforced thermoplastics with an inelastically-informed strategy},
  author={Dey, Argha Protim and Welschinger, Fabian and Schneider, Matti and Gajek, Sebastian and B{\"o}hlke, Thomas},
  journal={Archive of Applied Mechanics},
  volume={92},
  number={9},
  pages={2733--2755},
  year={2022},
  publisher={Springer}
}

@article{kalidindi2022digital,
  title={Digital twins for materials},
  author={Kalidindi, Surya R and Buzzy, Michael and Boyce, Brad L and Dingreville, Remi},
  journal={Frontiers in Materials},
  volume={9},
  pages={818535},
  year={2022},
  publisher={Frontiers Media SA}
}

@article{tikarrouchine2018three,
  title={Three-dimensional FE2 method for the simulation of non-linear, rate-dependent response of composite structures},
  author={Tikarrouchine, E and Chatzigeorgiou, George and Praud, Francis and Piotrowski, Boris and Chemisky, Yves and Meraghni, Fodil},
  journal={Composite Structures},
  volume={193},
  pages={165--179},
  year={2018},
  publisher={Elsevier}
}

@article{dressler2019heterogeneities,
  title={Heterogeneities dominate mechanical performance of additively manufactured metal lattice struts},
  author={Dressler, Amber D and Jost, Elliott W and Miers, John C and Moore, David G and Seepersad, Carolyn C and Boyce, Brad L},
  journal={Additive Manufacturing},
  volume={28},
  pages={692--703},
  year={2019},
  publisher={Elsevier}
}

@article{roach2020size,
  title={Size-dependent stochastic tensile properties in additively manufactured 316L stainless steel},
  author={Roach, Ashley M and White, Benjamin C and Garland, Anthony and Jared, Bradley H and Carroll, Jay D and Boyce, Brad L},
  journal={Additive Manufacturing},
  volume={32},
  pages={101090},
  year={2020},
  publisher={Elsevier}
}

@article{karthik2021heterogeneous,
  title={Heterogeneous aspects of additive manufactured metallic parts: a review},
  author={Karthik, GM and Kim, Hyoung Seop},
  journal={Metals and Materials International},
  volume={27},
  number={1},
  pages={1--39},
  year={2021},
  publisher={Springer}
}

@article{hu2017uncertainty,
  title={Uncertainty quantification and management in additive manufacturing: current status, needs, and opportunities},
  author={Hu, Zhen and Mahadevan, Sankaran},
  journal={The International Journal of Advanced Manufacturing Technology},
  volume={93},
  number={5},
  pages={2855--2874},
  year={2017},
  publisher={Springer}
}

@article{godfrey2022heterogeneity,
  title={Heterogeneity and solidification pathways in additively manufactured 316L stainless steels},
  author={Godfrey, Amy J and Simpson, J and Leonard, D and Sisco, K and Dehoff, RR and Babu, SS},
  journal={Metallurgical and Materials Transactions A},
  volume={53},
  number={9},
  pages={3321--3340},
  year={2022},
  publisher={Springer}
}

@book{torquato,
author = "S. Torquato",
title = "Random Heterogeneous Materials",
publisher = "Springer",
address = "New York, NY",
year = "2002"
}

@article{heckman2020automated,
  title={Automated high-throughput tensile testing reveals stochastic process parameter sensitivity},
  author={Heckman, Nathan M and Ivanoff, Thomas A and Roach, Ashley M and Jared, Bradley H and Tung, Daniel J and Brown-Shaklee, Harlan J and Huber, Todd and Saiz, David J and Koepke, Josh R and Rodelas, Jeffrey M and others},
  journal={Materials Science and Engineering: A},
  volume={772},
  pages={138632},
  year={2020},
  publisher={Elsevier}
}

@article{salzbrenner2017high,
  title={High-throughput stochastic tensile performance of additively manufactured stainless steel},
  author={Salzbrenner, Bradley C and Rodelas, Jeffrey M and Madison, Jonathan D and Jared, Bradley H and Swiler, Laura P and Shen, Yu-Lin and Boyce, Brad L},
  journal={Journal of Materials Processing Technology},
  volume={241},
  pages={1--12},
  year={2017},
  publisher={Elsevier}
}

@article{fernandez2022creep,
  title={Creep anisotropy modeling and uncertainty quantification of an additively manufactured Ni-based superalloy},
  author={Fernandez-Zelaia, Patxi and Lee, Yousub and Dryepondt, Sebastien and Kirka, Michael M},
  journal={International Journal of plasticity},
  volume={151},
  pages={103177},
  year={2022},
  publisher={Elsevier}
}

@article{muir2021damage,
  title={Damage mechanism identification in composites via machine learning and acoustic emission},
  author={Muir, C and Swaminathan, B and Almansour, AS and Sevener, K and Smith, C and Presby, M and Kiser, JD and Pollock, TM and Daly, S},
  journal={npj Computational Materials},
  volume={7},
  number={1},
  pages={95},
  year={2021},
  publisher={Nature Publishing Group UK London}
}

@article{papamakarios2021normalizing,
  title={Normalizing flows for probabilistic modeling and inference},
  author={Papamakarios, George and Nalisnick, Eric and Rezende, Danilo Jimenez and Mohamed, Shakir and Lakshminarayanan, Balaji},
  journal={Journal of Machine Learning Research},
  volume={22},
  number={57},
  pages={1--64},
  year={2021}
}

@article{kennedy2001bayesian,
  title={Bayesian calibration of computer models},
  author={Kennedy, Marc C and O'Hagan, Anthony},
  journal={Journal of the Royal Statistical Society: Series B (Statistical Methodology)},
  volume={63},
  number={3},
  pages={425--464},
  year={2001},
  publisher={Wiley Online Library}
}

@article{generale2024inverse,
  title={Inverse stochastic microstructure design},
  author={Generale, Adam P and Robertson, Andreas E and Kelly, Conlain and Kalidindi, Surya R},
  journal={Acta Materialia},
  volume={271},
  pages={119877},
  year={2024},
  publisher={Elsevier}
}

@article{zang2025psp,
  title={PSP-GEN: Stochastic inversion of the Process--Structure--Property chain in materials design through deep, generative probabilistic modeling},
  author={Zang, Yaohua and Koutsourelakis, Phaedon-Stelios},
  journal={Acta Materialia},
  volume={284},
  pages={120600},
  year={2025},
  publisher={Elsevier}
}

@article{venkatraman2022bayesian,
  title={Bayesian analysis of parametric uncertainties and model form probabilities for two different crystal plasticity models of lamellar grains in $\alpha$+ $\beta$ titanium alloys},
  author={Venkatraman, Aditya and McDowell, David L and Kalidindi, Surya R},
  journal={International Journal of Plasticity},
  volume={154},
  pages={103289},
  year={2022},
  publisher={Elsevier}
}

@article{khatamsaz2023gpr,
  title={Bayesian optimization with active learning of design constraints using an entropy-based approach},
  author={Khatamsaz, Danial and Vela, Brent and Singh, Prashant and Johnson, Duane D and Allaire, Douglas and Arr{\'o}yave, Raymundo},
  journal={npj Computational Materials},
  volume={9},
  number={1},
  pages={49},
  year={2023},
  publisher={Nature Publishing Group UK London}
}

@article{pasparakis2025bayesian,
  title={Bayesian neural networks for predicting uncertainty in full-field material response},
  author={Pasparakis, George D and Graham-Brady, Lori and Shields, Michael D},
  journal={Computer Methods in Applied Mechanics and Engineering},
  volume={433},
  pages={117486},
  year={2025},
  publisher={Elsevier}
}

@article{fullwoodsurvey,
author = "D.T. Fullwood and S.R. Niezgoda and B.L. Adams and S.R. Kalidindi",
title = "Microstructure Sensitive Design for Performance Optimization",
journal = "Prog. Mater. Sci.",
volume = "55",
pages = "477-562",
year = "2010",
doi="10.1016/j.pmatsci.2009.08.002"
}

@article{brough_processstructure_phasefield,
title = {Microstructure-based knowledge systems for capturing process-structure evolution linkages},
journal = {Current Opinion in Solid State and Materials Science},
volume = {21},
pages = {129-140},
year = {2017},
doi = {https://doi.org/10.1016/j.cossms.2016.05.002},
author = {David B. Brough and Daniel Wheeler and James A. Warren and Surya R. Kalidindi},
}

@article{chan2024hyperdiffusion,
  title={Estimating epistemic and aleatoric uncertainty with a single model},
  author={Chan, Matthew and Molina, Maria and Metzler, Chris},
  journal={Advances in Neural Information Processing Systems},
  volume={37},
  pages={109845--109870},
  year={2024}
}

@article{pourkamali2024probabilistic,
  title={Probabilistic Neural Networks (PNNs) for Modeling Aleatoric Uncertainty in Scientific Machine Learning},
  author={Pourkamali-Anaraki, Farhad and Husseini, Jamal F and Stapleton, Scott E},
  journal={IEEE Access},
  year={2024},
  publisher={IEEE}
}

@article{sun2022alpha,
  title={$\alpha$-deep probabilistic inference ($\alpha$-dpi): efficient uncertainty quantification from exoplanet astrometry to black hole feature extraction},
  author={Sun, He and Bouman, Katherine L and Tiede, Paul and Wang, Jason J and Blunt, Sarah and Mawet, Dimitri},
  journal={The Astrophysical Journal},
  volume={932},
  number={2},
  pages={99},
  year={2022},
  publisher={IOP Publishing}
}

@article{neal1992bayesian,
  title={Bayesian learning via stochastic dynamics},
  author={Neal, Radford},
  journal={Advances in neural information processing systems},
  volume={5},
  year={1992}
}

@inproceedings{daubener2025elbo,
  title={ELBO, regularized maximum likelihood, and their common one-sample approximation for training stochastic neural networks},
  author={D{\"a}ubener, Sina and Damm, Simon and Fischer, Asja},
  booktitle={The 41st Conference on Uncertainty in Artificial Intelligence},
  year={2025},
}

@inproceedings{li2015generative,
  title={Generative moment matching networks},
  author={Li, Yujia and Swersky, Kevin and Zemel, Rich},
  booktitle={International conference on machine learning},
  pages={1718--1727},
  year={2015},
  organization={PMLR}
}

@article{borgwardt2006integrating,
  title={Integrating structured biological data by kernel maximum mean discrepancy},
  author={Borgwardt, Karsten M and Gretton, Arthur and Rasch, Malte J and Kriegel, Hans-Peter and Sch{\"o}lkopf, Bernhard and Smola, Alex J},
  journal={Bioinformatics},
  volume={22},
  number={14},
  pages={e49--e57},
  year={2006},
  publisher={Oxford University Press}
}

@inproceedings{generale2023bayesian,
  title={A bayesian approach to designing microstructures and processing pathways for tailored material properties},
  author={Generale, Adam P and Kelly, Conlain and Harrington, Grayson and Robertson, Andreas Euan and Buzzy, Michael and Kalidindi, Surya},
  booktitle={AI for Accelerated Materials Design-NeurIPS 2023 Workshop},
  year={2023}
}

@article{cui2020calibrated,
  title={Calibrated reliable regression using maximum mean discrepancy},
  author={Cui, Peng and Hu, Wenbo and Zhu, Jun},
  journal={Advances in Neural Information Processing Systems},
  volume={33},
  pages={17164--17175},
  year={2020}
}

@Misc{functorch2021,
  author =       {Horace He, Richard Zou},
  title =        {functorch: JAX-like composable function transforms for PyTorch},
  howpublished = {\url{https://github.com/pytorch/functorch}},
  year =         {2021}
}

@article{venkatraman2025matse,
  title={MATSE-Multi-Fidelity Augmented Time-Series Emulation: Galvanic Corrosion Applications},
  author={Venkatraman, Aditya and Katona, Ryan and Montes de Oca Zapiain, David},
  journal={Machine Learning: Science and Technology},
  year={2025}
}

@article{girard2002gaussian,
  title={Gaussian process priors with uncertain inputs application to multiple-step ahead time series forecasting},
  author={Girard, Agathe and Rasmussen, Carl and Candela, Joaquin Q and Murray-Smith, Roderick},
  journal={Advances in neural information processing systems},
  volume={15},
  year={2002}
}

@article{calinon2020reimanniangaussians,
  title={Gaussians on Riemannian manifolds: Applications for robot learning and adaptive control},
  author={Calinon, Sylvain},
  journal={IEEE Robotics \& Automation Magazine},
  volume={27},
  number={2},
  pages={33--45},
  year={2020},
  publisher={IEEE}
}

@article{pennec2006intrinsicreimannian,
  title={Intrinsic statistics on Riemannian manifolds: Basic tools for geometric measurements},
  author={Pennec, Xavier},
  journal={Journal of Mathematical Imaging and Vision},
  volume={25},
  pages={127--154},
  year={2006},
  publisher={Springer}
}

@article{sailynoja2022graphical,
  title={Graphical test for discrete uniformity and its applications in goodness-of-fit evaluation and multiple sample comparison},
  author={S{\"a}ilynoja, Teemu and B{\"u}rkner, Paul-Christian and Vehtari, Aki},
  journal={Statistics and Computing},
  volume={32},
  number={2},
  pages={32},
  year={2022},
  publisher={Springer}
}

@article{gruich2023clarifying,
  title={Clarifying trust of materials property predictions using neural networks with distribution-specific uncertainty quantification},
  author={Gruich, Cameron J and Madhavan, Varun and Wang, Yixin and Goldsmith, Bryan R},
  journal={Machine Learning: Science and Technology},
  volume={4},
  number={2},
  pages={025019},
  year={2023},
  publisher={IOP Publishing}
}

@software{skopt,
  author = {Head, Tim and MechCoder and Louppe, Gilles and Shcherbatyi, Iaroslav and fcharras and Vinícius, Zé and cmmalone and Schr\"oder, Christopher and nel215 and Campos, Nuno and Young, Todd and Cereda, Stefano and Fan, Thomas and rene-rex and Shi, Kejia (KJ) and Schwabedal, Justus and carlosdanielcsantos and Hvass-Labs and Pak, Mikhail and SoManyUsernamesTaken and Callaway, Fred and Est\'eve, Lo\"ic and Besson, Lilian and Cherti, Mehdi and Pfannschmidt, Karlson and Linzberger, Fabian and Cauet, Christophe and Gut, Anna and Mueller, Andreas and Fabisch, Alexander},
  title = {scikit-optimize},
  url = {https://doi.org/10.5281/zenodo.1207017},
  doi = {10.5281/zenodo.1207017},
  version = {v0.5.2},
  date = {2018},
}

@article{kennard1969maxmin,
  title={Computer aided design of experiments},
  author={Kennard, Ronald W and Stone, Larry A},
  journal={Technometrics},
  volume={11},
  number={1},
  pages={137--148},
  year={1969},
  publisher={Taylor \& Francis}
}

@article{mak2018maxmin,
  title={Minimax and minimax projection designs using clustering},
  author={Mak, Simon and Joseph, V Roshan},
  journal={Journal of Computational and Graphical Statistics},
  volume={27},
  number={1},
  pages={166--178},
  year={2018},
  publisher={Taylor \& Francis}
}

@article{robertson2022efficient,
  title={Efficient generation of anisotropic N-field microstructures from 2-point statistics using multi-output Gaussian random fields},
  author={Robertson, Andreas E and Kalidindi, Surya R},
  journal={Acta Materialia},
  volume={232},
  pages={117927},
  year={2022},
  publisher={Elsevier}
}

@article{dingreville2024benchmarking,
  title={Benchmarking machine learning strategies for phase-field problems},
  author={Dingreville, R{\'e}mi and Roberston, Andreas E and Attari, Vahid and Greenwood, Michael and Ofori-Opoku, Nana and Ramesh, Mythreyi and Voorhees, Peter W and Zhang, Qian},
  journal={Modelling and Simulation in Materials Science and Engineering},
  volume={32},
  number={6},
  pages={065019},
  year={2024},
  publisher={IOP Publishing}
}

@article{loshchilov2016sgdr,
  title={Sgdr: Stochastic gradient descent with warm restarts},
  author={Loshchilov, Ilya and Hutter, Frank},
  journal={arXiv preprint arXiv:1608.03983},
  year={2016}
}

@article{kingma2014adam,
  title={Adam: A method for stochastic optimization},
  author={Kingma, Diederik P and Ba, Jimmy},
  journal={arXiv preprint arXiv:1412.6980},
  year={2014}
}

@article{reddi2019convergence,
  title={On the convergence of adam and beyond},
  author={Reddi, Sashank J and Kale, Satyen and Kumar, Sanjiv},
  journal={arXiv preprint arXiv:1904.09237},
  year={2019}
}

@article{lebensohn2013modeling,
  title={Modeling void growth in polycrystalline materials},
  author={Lebensohn, Ricardo A and Escobedo, Juan P and Cerreta, Ellen K and Dennis-Koller, Darcie and Bronkhorst, Curt A and Bingert, John F},
  journal={Acta Materialia},
  volume={61},
  number={18},
  pages={6918--6932},
  year={2013},
  publisher={Elsevier}
}

@article{boyce2023machine,
  title={Machine learning for materials science: Barriers to broader adoption},
  author={Boyce, Brad and Dingreville, Remi and Desai, Saaketh and Walker, Elise and Shilt, Troy and Bassett, Kimberly L and Wixom, Ryan R and Stebner, Aaron P and Arroyave, Raymundo and Hattrick-Simpers, Jason and others},
  journal={Matter},
  volume={6},
  number={5},
  pages={1320--1323},
  year={2023},
  publisher={Elsevier}
}

@article{walker2024unsupervised,
  title={Unsupervised physics-informed disentanglement of multimodal data},
  author={Walker, Elise and Trask, Nathaniel and Martinez, Carianne and Lee, Kookjin and Actor, Jonas A and Saha, Sourav and Shilt, Troy and Vizoso, Daniel and Dingreville, Remi and Boyce, Brad L},
  journal={Foundations of Data Science},
  volume={7},
  year={2024},
}

@article{desai2022learning,
  title={Learning time-dependent deposition protocols to design thin films via genetic algorithms},
  author={Desai, Saaketh and Dingreville, R{\'e}mi},
  journal={Materials \& Design},
  volume={219},
  pages={110815},
  year={2022},
  publisher={Elsevier}
}

@article{szymanski2023autonomous,
  title={An autonomous laboratory for the accelerated synthesis of novel materials},
  author={Szymanski, Nathan J and Rendy, Bernardus and Fei, Yuxing and Kumar, Rishi E and He, Tanjin and Milsted, David and McDermott, Matthew J and Gallant, Max and Cubuk, Ekin Dogus and Merchant, Amil and others},
  journal={Nature},
  volume={624},
  number={7990},
  pages={86--91},
  year={2023},
  publisher={Nature Publishing Group UK London}
}

@article{clark1994modelling,
  title={Modelling the effective conductivity function of an arbitrary two--dimensional polycrystal using sequential laminates},
  author={Clark, Karen E and Milton, Graeme W},
  journal={Proceedings of the Royal Society of Edinburgh Section A: Mathematics},
  volume={124},
  number={4},
  pages={757--783},
  year={1994},
  publisher={Royal Society of Edinburgh Scotland Foundation}
}

@book{milton2022theory,
  title={The Theory of Composites},
  author={Milton, Graeme W},
  year={2022},
  publisher={SIAM}
}

\end{document}